\documentclass{article}
\DeclareUnicodeCharacter{0279}{\textipa{r}}

\usepackage{PRIMEarxiv}

\usepackage[utf8]{inputenc} 
\usepackage[T1]{fontenc}    
\usepackage{hyperref}       
\usepackage{url}            
\usepackage{booktabs}       
\usepackage{amsfonts}       
\usepackage{nicefrac}       
\usepackage{microtype}      
\usepackage{lipsum}
\usepackage{fancyhdr}       
\usepackage{graphicx}       
\usepackage{tabularx}
\usepackage{tipa}
\DeclareFontSubstitution{T3}{ptm}{m}{n}
\usepackage{caption}
\usepackage{pdflscape}
\usepackage{ulem}
\usepackage[
    backend=biber,
    natbib=true,
    style=apa,
]{biblatex}
\DeclareFieldFormat{journaltitle}{\mkbibemph{#1}}
\usepackage{xcolor}
\usepackage[title]{appendix}
\graphicspath{{media/}}     

\title{Automatic Speech Recognition for Non-Native English: Accuracy and Disfluency Handling
\thanks{\textit{\underline{Citation}}: 
\textbf{McGuire, M. (2025). Automatic speech recognition for non-native English: Accuracy and disfluency handling. arXiv. https://doi.org/10.48550/arXiv.2503.06924}} 
}

\author{
  Michael McGuire \\
  Department of English \\
  Doshisha University \\
  Kyoto, Japan\\
  mmcguire [at] mail.doshisha.ac.jp
}

\begin{document}
\maketitle

\begin{abstract}
Automatic speech recognition (ASR) has been an essential component of computer assisted language learning (CALL) and computer assisted language testing (CALT) for many years. As this technology continues to develop rapidly, it is important to evaluate the accuracy of current ASR systems for language learning applications. This study assesses five cutting-edge ASR systems' recognition of non-native accented English speech using recordings from the L2-ARCTIC corpus, featuring speakers from six different L1 backgrounds (Arabic, Chinese, Hindi, Korean, Spanish, and Vietnamese), in the form of both read and spontaneous speech. The read speech consisted of 2,400 single sentence recordings from 24 speakers, while the spontaneous speech included narrative recordings from 22 speakers. Results showed that for read speech, Whisper and AssemblyAI achieved the best accuracy with mean Match Error Rates (\textit{MER}) of 0.054 and 0.056 respectively, approaching human-level accuracy. For spontaneous speech, RevAI performed best with a mean \textit{MER} of 0.063. The study also examined how each system handled disfluencies such as filler words, repetitions, and revisions, finding significant variation in performance across systems and disfluency types. While processing speed varied considerably between systems, longer processing times did not necessarily correlate with better accuracy. By detailing the performance of several of the most recent, widely-available ASR systems on non-native English speech, this study aims to help language instructors and researchers understand the strengths and weaknesses of each system and identify which may be suitable for specific use cases.
\end{abstract}

\keywords{automatic speech recognition \and learner English \and non-native speech \and computer assisted language learning}

\section{Introduction}

Automatic speech recognition (ASR), or the automated conversion of spoken language into text, has become commonplace with the development of computers, smartphones, and home automation. Many people now talk to their devices rather than type. The use cases for ASR go far beyond consumer products, however. This technology has had an effect on language learning, and has been considered an essential component of computer assisted language learning (CALL) and computer assisted language testing (CALT) for many years \citep{ehsani_speech_1998}.

ASR can be applied to CALL and CALT in numerous ways. \citet{shadiev_review_2023} reviewed 26 articles on the use of ASR for language learning published between 2014 and 2020 and found that the most common use was for pronunciation skills, followed by listening, writing, communication, grammar, word recognition, and vocabulary (p. 78). Studies that have looked specifically at pronunciation practice have found that ASR can be used to detect differences in phonemes, words, intonation, stress, and other features. This can be used to give learners explicit corrective feedback, which is often absent in other forms of pronunciation practice \citep{mccrocklin_asr-based_2019} and can greatly help pronunciation development \citep{saito_effects_2012}. A recent meta-analysis \citep{thi-nhu_ngo_effectiveness_2024} on the effectiveness of ASR in improving ESL/EFL pronunciation found an overall medium effect size (Hedge’s $g = 0.69$) across 15 studies, and a large effect size ($g = 0.86$) in eight studies incorporating explicit corrective feedback.

ASR can also be applied to both proficiency testing and progress monitoring. It is used in the speaking sections of many standardized language proficiency online tests such as the Test of English for International Communication (TOEIC), the Test of English as a Foreign Language Internet-based test (TOEFL iBT), and the International English Language Testing System (IELTS). Other forms of oral proficiency assessment such as elicited imitation tests can be automated using ASR \citep{mcguire_assessing_2025}. Some projects combine ASR with large language models (LLMs) to create interactive AI characters for conversation practice and proficiency assessment, such as the InteLLA project \citep{saeki_intella_2024}. Other uses include fluency analysis, grammar pattern drills, and read aloud practice. ASR can make assessment easier for teachers by transcribing student presentations or discussions.

Researchers have examined the use of ASR for language learning over the past several decades, but the technology is rapidly developing and has reached its highest levels of accuracy in the past few years thanks to advances in neural networks and transformer systems. As a result, studies quickly become outdated as the technology advances and new ASR models are created. While the present study will certainly meet the same fate, it is worthwhile to examine where the technology stands today. This study assesses five cutting-edge ASR systems’ accuracy in the recognition of non-native accented English speech from six different L1 backgrounds in the form of both read and spontaneous speech, as well as how verbatim each system transcribes disfluencies in spontaneous speech. By detailing the performance of several of the most recent, widely-available ASR systems on non-native English speech, this study aims to help language instructors and researchers understand the strengths and weaknesses of each system and identify which may be suitable for specific use cases.

\subsection{Automatic Speech Recognition (ASR)}

Automatic speech recognition (ASR) is any kind of technology–primarily software–that converts speech into text. Rather than transcribing by hand, ASR allows the task to be completed automatically by a computer. ASR has beshadievcome increasingly common in consumer products for voice dictation, search, home automation, translation, and a variety of other tasks.

\subsubsection{Development of ASR}

Early ASR recognized single words in isolation. The first working ASR system was developed in the 1950s at Bell Labs in the United States which could recognize spoken numbers \citep{davis_automatic_1952}. In the 1960s, researchers in Japan developed systems that could recognize different vowels \citep{suzuki_recognition_1961} and phonemes \citep{sakai_phonetic_1961}.

It wasn’t until the late 1980s that large vocabulary continuous speech recognition (LVCSR) became possible with the groundbreaking SPHINX ASR system \citep{lee_large-vocabulary_1988}, which is still maintained by Carnegie Mellon University (CMU) as an open-source project \citep{cmusphinx_2025}. This system used a design called a hidden Markov model (HMM) which could calculate the probability of the next sound based on statistics collected from supervised training data i.e. audio paired with transcriptions \citep{rabiner_tutorial_1989}. These statistical values were stored as parameters, and advancements in computer technology would allow for systems with greater numbers of parameters.

HMM based ASR was improved and refined through the 1990s and 2000s, but was not surpassed until the advent of the Deep Neural Network (DNN) approach in the 2010s \citep{hinton_deep_2012}. Rather than being purely statistics based like the HMM, a DNN was a layered network that was created through exposure to a much larger amount training data, and thus had a much greater number of parameters. ASR systems that combined HMM and DNN were called Hybrid systems, which are still used today \citep{steadman_ursa_2024}. Hybrid systems recognized speech by combining separate models that were trained separately to do different tasks \citep{jaitly_application_2012}. For example, an acoustic model was trained on audio to convert speech into phonemes, a pronunciation model, usually based on a pronunciation dictionary, combined the identified phonemes into words, and a language model trained on text predicted possible next words. These different models operated in parallel for greater accuracy and efficiency.

With each advancement, the amount of training data required grew. However, the amount of supervised training data–speech audio transcribed by humans–was limited, and it was expensive and time-consuming to create more. One advantage to having a separate acoustic model was that it could be pre-trained with large quantities of much more readily available unsupervised data–speech audio without transcriptions. The acoustic model could learn speech features from pre-training, which then made the training on supervised data more efficient \citep{dahl_context-dependent_2012}.

Around the same time, a different approach called end-to-end was proposed \citep{graves_sequence_2012}. Instead of using separate models like in Hybrid systems, an end-to-end system combined them all into a single neural network. Doing this required a massive amount of training data, however. The end-to-end system introduced by \citet{graves_sequence_2012} was based on a Recurrent Neural Network (RNN), which was a type of DNN designed specifically for continuous sequential data like speech. It was recurrent because it constantly looped back to previous inputs when generating output, which gave it a memory of what came before. The RNN architecture introduced by \citet{graves_sequence_2012}, the RNN-Transducer (RNN-T) was one of the first end-to-end systems that could perform real-time streaming ASR, i.e. output text while still receiving input, as opposed to asynchronous ASR which required a completed input before generating output.

Next, attention-based RNN systems proposed by \citet{chorowski_end--end_2014} adapted a novel approach from machine translation \citep{bahdanau_neural_2014} and applied it to ASR. In an attention-based system, the separate model approach from hybrid systems was replaced with an encoder-decoder structure. The encoder and decoder were both RNNs that were connected; the encoder converted audio into features, and the decoder used those features to predict the output text sequence. The encoder and decoder were connected by an attention mechanism. While normal RNNs handled sequential data in order while constantly looping back to review everything in their memory, an attention mechanism weighed and selectively chose which features from the encoder were relevant to the current state of the decoder. These attention-based RNN systems could not do streaming ASR, but they were more generally more accurate than RNN-Ts, especially with longer, more complex input. In 2015, researchers behind the Deep Speech 2 ASR system reported accuracy matching professional human transcription in both English and Mandarin \citep{amodei_deep_2015}. \citet{chan_listen_2016} refined the attention-based system, improving accuracy and making it more practical to train and use.

The next major breakthrough, a new type of neural network called a transformer, was introduced in the seminal paper ``Attention is All You Need'' \citep{vaswani_attention_2017}. As the title suggests, transformers replaced RNNs with attention mechanisms, including a new form of attention mechanism called self-attention. Self-attention allowed each part of a sequence to pay attention to all the other parts of the same sequence. Transformers combined multiple layers of self-attention: input layers where the input attended to itself in the encoder, output layers where the output attended to itself in the decoder, and cross-attenion layers where the decoder attended to the encoder output. These layers allowed transformers to handle context in more complex sequences, recognize more complex patterns and long-range dependencies, and process the information in more sophisticated ways. Transformers were also the basis of the generative large-language models that are popular today such as the GPT series \citep{radford_improving_2018}.

The ASR systems evaluated in this study differ in their approaches, architectures, as well as the amount and type of training data and parameters. However, because many are proprietary systems, not all of this information is disclosed. Please see Subsection~\ref{subsec:asr_systems} for specific details for each system.

\subsubsection{Multilingual ASR and Accented Speech}

The early development of ASR was language-specific; but in the late 90s there was discussion of creating multilingual ASR systems that could transcribe less common languages without providing training data from that language \citep{byrne_towards_2000}. Creating a language-universal ASR system could be achieved by using training data from many different languages that cover the universal phone set (UPS) \citep{lin_study_2009}. However, even with multilingual ASR, the training data typically came from native speakers of each language, and rarely from non-native speakers. Thus, an important question that this study investigates is how well can multilingual ASR systems handle non-native accented speech.

Accented speech, whether non-native speech or regional accents or dialects, has been considered an important issue for ASR and machine translation for many years \citep{neri_automatic_2003}. As numerous studies throughout the development of ASR have shown, ASR transcriptions of accented speech receive higher error rates than non-accented speech \citep{coniam_voice_1999,derwing_does_2000,graham_evaluating_2024,hirai_speech--text_2024,koenecke_racial_2020}. It has been found, however, that using diverse multilingual training data \citep{vergyri_automatic_2010} and accent adaptive algorithms \citep{bell_adaptation_2021} can greatly improve the recognition of accented speech. 

The reduced accuracy of ASR systems on accented speech is likely related to the sourcing of accurate supervised training data, which is much more difficult and expensive for accented speech than for non-accented speech \citep{hinsvark_accented_2021}. One creative solution proposed recently is to use accented speech recordings to create accented text to speech models, i.e. computer generated speech with accents, which could be used to generate training data for ASR \citep{do_improving_2024}. Presently, more research attention is focusing on improving the recognition of accented speech, and it is one strand in the theme of the 2025 Interspeech Conference on speech processing technology. This study aims to contribute to this push by assessing the performance of five of the most recent, cutting-edge, publicly available ASR systems.

\subsection{Measuring ASR Accuracy}

To measure the accuracy of ASR, the hypothesis–the transcribed text–is compared to the reference– the original prompt or human transcription, also called the ground truth. The hypothesis and reference are aligned to see if words are the same, different, or in the wrong position. If the first word in the hypothesis is ``the'', and the first word in the reference is ``the'', this is considered a hit ($H$), or a correct word. Errors include substitutions ($S$), deletions ($D$), and insertions ($I$). If the reference is ``please open the windows'' and the hypothesis is ``open a window'', there is one hit open, two substitutions \textit{the} to \textit{a} and \textit{windows} to \textit{window}, and one deletion \textit{please}. The number of hits, substitutions, deletions, and insertions is then used to calculate an error rate. 

The most commonly used error rate for ASR is called Word Error Rate (\textit{WER}), which divides the sum of substitutions, deletions, and insertions by the number of words in the reference counted as the sum of substitutions, deletions, and hits. \textit{WER} is computed as
\begin{equation}
WER = \frac{S + D + I}{S + D + H}
\end{equation}
This gives the ratio of the number of errors to the number of words in the reference, which is generally between 0 and 1. However, \textit{WER} will exceed 1 if the hypothesis contains many insertions, which will result in excess penalty to the accuracy score. 
	
A critical paper by \citet{morris_wer_2004} argued that Match Error Rate (\textit{MER}) is a much more appropriate metric for most speech recognition applications. While \textit{WER} weighs errors relative to the length of the reference, \textit{MER} weighs errors relative to the total alignment count by including insertions in the denominator and thus represents the probability of any word being incorrect. \textit{MER} divides the number of errors by the total number of alignment decisions and is computed as
\begin{equation}
MER = \frac{S + D + I}{S + D + H + I}
\end{equation}
The difference is small, but important. Ultimately, \textit{WER} measures the edit cost, or how many corrections are needed to make the hypothesis match the reference. \textit{MER}, on the other hand, measures how much of the original message is successfully communicated. To illustrate, \citet{morris_wer_2004} describes two different systems: 
\begin{quote}
Consider an ASR system for telephone information retrieval which outputs a wrong word for each word input, so that $S=N_1, H=D=I=0$ and $WER = N_1/N_1 = 100\%$. A similar system which outputs two wrong words for each input word has $S=N_1, H=D=0, I = N_1$, and $WER = (N_1+N_1)/N_1 = 200\%$. However, both systems communicate exactly zero information and have zero utility, so the performance difference suggested by relative \textit{WER} scores is, in this case, simply misleading. (p. 1)
\end{quote}
The \textit{WER} of the second system is 2 (200\%) because of all the edits required to fix the mistakes, and this only matters when you need to actually make the edits by hand. The \textit{MER} for both systems, however, would be 1 (100\%) because they both fail at communicating any of the information in the reference. When taking the mean of scores, any \textit{WER} score above 1 will distort other scores. \textit{MER} solves this problem by always being between 0 and 1, and makes more sense for most applications.

\subsection{Human Accuracy}

The goalpost for ASR development has always been to achieve human-level accuracy, but what does that mean? Accuracy is judged by comparing ASR transcriptions to human transcriptions, but human transcriptions are not necessarily perfect. People may disagree on what they hear, and they may have different transcription conventions even when they do agree. Conversational speech is intrinsically ambiguous due to the shared context and background knowledge between speakers that transcribers do not have access to \citep{stolcke_comparing_2017}. Accented and non-native speech may be even more challenging for human transcribers due to variations in pronunciation. Therefore, human transcription will also have a degree of error.

As mentioned earlier, \citet{amodei_deep_2015} claimed that the Deep Speech 2 model was competitive with human transcribers on read English speech. They compared both ASR transcriptions and crowdsourced human transcriptions of several data sets, including two sets of read Wall Street Journal news articles and the LibriSpeech \citep{panayotov_librispeech_2015} audiobook corpus. Because these corpora are edited read speech, the ground truth should be completely accurate. The crowdsourced human transcriptions had \textit{WER}s of 5.03 and 8.08 on the two Wall Street Journal data sets and 5.83 on LibriSpeech, while their Deep Speech 2 ASR transcriptions had \textit{WER}s of 3.60, 4.98, and 5.33 respectively. 

Unlike read speech, conversational speech does not have an objective ground truth outside of human transcription. \citet{xiong_achieving_2017} was the first study to claim human parity in conversational speech with an ASR system. This study used portions of the National Institute of Standards and Technology (NIST) 2000 conversational telephone speech data set \citep{przybocki_2000_2001}. They scored both their ASR transcriptions and professional two-pass human transcriptions (one person transcribed and a second person checked for errors) against the ground truth; however, the ground truth transcriptions of this data set were also done by human transcribers \citep{fiscus_2000_2000}, making it somewhat circular. Nevertheless, this cannot be avoided with conversational speech, and the NIST 2000 data set has been used in many studies. \citet{xiong_achieving_2017} found the error rate of their human transcriptions to be 11.3\% on one portion of the data and 5.9\% on another, while their ASR system achieved 11\% and 5.8\% respectively.

These studies help quantify human transcription accuracy and serve as a reference point for testing ASR systems. Following the results of these studies, the error rate of human transcribers for both read and spontaneous speech can be as low as 5.8\%, which has been used as a benchmark in other studies \citep{radford_robust_2022}.

\subsection{Disfluencies and Disfluency Handling in ASR}
Disfluencies are both verbal and non-verbal hesitations that interrupt the flow of speech and are common in spontaneous or conversational speech \citep{redford_fluency_2015}. They may take the form of pauses, filler words such as ``um'' or ``uh'', repetitions such as ``the the'', or revisions such as ``he went I mean she went''. Disfluencies generally occur when a speaker struggles with planning speech \citep{derwing_fluency_2022}, so they are common in L2 speech. 

Disfluencies are not always transcribed accurately by ASR systems, which are often designed to omit them in order to improve the readability of their output \citep{amann_augmenting_2024}. Such omission may be acceptable when only the primary message is required, but it can be problematic when verbatim transcription is needed. Furthermore, the omission of disfluencies will greatly impact ASR transcription accuracy measurements if compared to a ground truth which retains them, such as for the spontaneous speech samples used in this study. To address this, many ASR APIs now include options to allow the user to decide whether to omit or retain disfluencies. These options are not perfect, however, so determining how well ASR systems handle disfluencies is important. This study looks specifically at this by comparing the transcriptions from each ASR system under both conditions–disfluencies omitted and disfluencies retained.

\section{Research Questions}

This study compares the performance of five widely-available, modern ASR systems on both read and spontaneous L2 English learner speech. The research questions are as follows:

\begin{quote}
\noindent\textbf{RQ1.} How do the selected automatic speech recognition systems compare in accuracy and processing time when transcribing L2 English speech in the form of (a) read speech and (b) spontaneous speech, and how do the accuracy rates vary by gender and L1 background?

\vspace{1em}

\noindent\textbf{RQ2.} How accurately do the ASR systems transcribe disfluencies in L2 speech?
\end{quote}

\section{Methods}

\subsection{Speech Samples}

The speech samples used to test the five ASR systems were taken from the L2-ARCTIC non-native English speech corpus \citep{zhao_l2-arctic_2018}. This corpus contains laboratory recordings and transcriptions of 24 speakers from six L1 backgrounds–Arabic, Chinese, Hindi, Korean, Spanish, and Vietnamese–with two male and two female speakers each. Each speaker was identified by a three to five letter combination such as ABA or MBMPS. The L2-ARCTIC corpus was chosen for its coverage of different L1s, balanced gender distribution, recording quality, transcriptions, and Creative Commons CC BY-NC 4.0 license. The corpus contains two sections: 26,867 elicited read-aloud single sentence recordings, and the suitcase corpus–recordings of an oral narrative task about a man and a woman mixing up their suitcases \citep{derwing_relationship_2009} from 22 of the 24 speakers. A transcription of each individual recording is also included.

\paragraph{RQ1a -- Read Speech.} Each speaker ($n = 24$) read aloud sentences from Carnegie Melon University's ARCTIC prompts \citep{kominek_cmu_2004}---around 1150 sentences between five to twenty words long extracted from out-of-copyright Project Gutenberg books. Both the corpus and the sentence list were designed to give complete coverage of American English phonemes for pronunciation research. Due to the sources, many of the sentence prompts used in the L2-ARCTIC speech corpus are far from typical L2 English learner speech, containing difficult character names (\textit{Whittemore}, \textit{MacDougall}, or \textit{Fitzburgh}), low-frequency words (\textit{acquiescence}, \textit{provocateur}, or \textit{immaculate}), or archaic expressions (\textit{gad}, \textit{by golly}, or \textit{doggone}). This study aims to assess the performances of ASR systems on more typical learner speech, so the full list of sentences was filtered to remove any sentences that contain words that do not appear in the New General Service List (NGSL) \citep{browne_new_2014}, a corpus-derived word list of the 2,800 most common words in general English. From the resulting filtered list, 100 sentences were then chosen randomly for inclusion in the analysis. The audio files for these sentences from each of the speakers were then collected, yielding 2,400 single sentence recordings ranging in length from 0.8 seconds to 7.9 seconds (averaging 2.98 seconds per recording) with a total duration of 7,142 seconds (119 minutes). In addition to the 24 speakers from the L2-ARCTIC corpus, recordings of the same sentences by four native speakers of American English taken from the CMU ARCTIC database \citep{kominek_cmu_2004} were used as a control for L1 effect analysis.

\paragraph{RQ1b and RQ2 -- Spontaneous Speech.} The suitcase corpus section of the L2-ARCTIC contains an oral narrative recording from each of $n = 22$ speakers, ranging from 27.5 to 235 seconds (averaging 71.8 seconds) with a total duration of 1566 seconds (about 26 minutes). These recordings provided high quality examples of longer-form spontaneous learner speech and include many disfluencies that are absent from the read speech samples. After processing all of the data once, several discrepancies were found in the ground truth transcriptions provided with the recordings such as inconsistent spellings and some missing words. These differences would negatively affect ASR transcription accuracy analysis, so all transcriptions were reviewed carefully by the author before being analyzed again. These revised transcriptions are available on the Open Science Foundation (OSF) page for this study (\url{https://osf.io/uszm9/}).

\subsection{ASR Systems}
\label{subsec:asr_systems}
Five of the newest ASR systems at the time of writing were chosen for this study: AssemblyAI's \texttt{Universal-2}, Deepgram's \texttt{Nova-2}, Rev.AI's \texttt{V2}, Speechmatics' \texttt{Ursa-2}, and Whisper \texttt{large-v3} by OpenAI. Only Whisper is open source and can be installed locally on your own computer; the other four systems are proprietary. This means that the exact details of their architectures are not always made public and their publications are for marketing and not peer reviewed. All of the information included in this study about their model architectures comes from their websites and marketing materials, in which they all claim to have developed the world’s most accurate ASR system. These proprietary ASR services may only be used through application programming interfaces (APIs) and cannot be installed locally. APIs allow you to remotely request transcription through the service provider’s servers. Thus, in order to compare the five systems in the same environment, all systems including Whisper were accessed through APIs using the Python programming language. 

The read speech samples used for RQ1a were processed by each system with default settings. However, because the spontaneous speech recordings used for RQ1b and RQ2 contain many disfluencies, they were processed through each ASR system twice: with and without disfluency retention settings enabled. This way, the performance of each ASR system can be compared under both conditions. The following is a brief introduction to each of these five ASR systems, and a summary of the technical details can be found in Table~\ref{tab:table1}.

\paragraph{AssemblyAI -- \texttt{Universal-2}.} (\url{https://www.assemblyai.com/}). AssemblyAI is a commercial ASR API service launched in 2017. Their \texttt{Universal-2} model \citep{universal2_2024} released in October, 2024 uses an RNN-T architecture with 600 million parameters and supports 20 languages. The encoder was pre-trained on 12.5 million hours of unsupervised multilingual audio and then the full model was fine-tuned on 300,000 hours of supervised audio. The API has a boolean (\texttt{true} or \texttt{false}) option called \texttt{disfluencies} for retaining or omitting filler words in the transcription. This option is \texttt{false} by default, so it must be set to \texttt{true} to retain filler words. 

\paragraph{Deepgram -- \texttt{Nova-2}.} (\url{https://deepgram.com/}). Deepgram is an AI company founded in 2015 that develops ASR, text-to-speech, and interactive voice AI services. Their Nova-2 model released in late 2023 uses an end-to-end transformer based architecture \citep{fox_introducing_2023} and supports 37 different languages. The number of parameters is not disclosed on their website, and the only information about the training data is that the previous model Nova-1 was trained on 47 billion tokens, i.e. words. The API has a boolean \texttt{filler\_words} option similar to AssemblyAI, which is \texttt{false} by default and must be set to \texttt{true} to retain filler words.

\paragraph{RevAI -- \texttt{V2}.} (\url{https://www.rev.ai/}). Rev started in 2010 as a human transcription and translation company, began researching ASR in 2016, and launched the RevAI ASR API in 2018. In 2022, they released their end-to-end model \texttt{V2}, which replaced the hybrid approach used in their previous \texttt{V1} model \citep{jette_what_2024,rev_press_rev_2022}. The architecture of \texttt{V2} is fairly unique; the encoder uses a modified transformer called a conformer \citep{gulati_conformer_2020}, and a connectionist temporal classification (CTC) \citep{graves_connectionist_2006} based decoder, and sends the results through a second pass attention-based model. Through their human transcription service, they have amassed their own collection of supervised training data of more than 3 million hours, though the amount actually used in training \texttt{V2} is not disclosed. The API supports 59 different languages, and has a boolean \texttt{remove\_disfluencies} option which can be adjusted to omit or retain filler words. When processing the read speech samples through Rev.AI, the API rejected recordings that were less than two seconds in length. These files needed to be padded with silence at the end to increase the length to above two seconds before they could be processed successfully. None of the other APIs had this issue. Additionally, RevAI inserts a speaker number and timestamp at the beginning of each transcript and sometimes adds atmospheric tags like \texttt{<silence>} and \texttt{<affirmative>}. Some API settings needed to be adjusted to remove these from transcriptions before \textit{MER} could be calculated.

\paragraph{Speechmatics -- \texttt{Ursa 2}.} (\url{https://www.speechmatics.com/}). Speechmatics was founded in 2006 by Dr. Tony Robinson, a researcher from Cambridge University who pioneered the use of neural networks in speech recognition \citep{robinson_recurrent_1991,lee_use_1996}. Their \texttt{Ursa 2} model \citep{steadman_ursa_2024} released in 2024 has a hybrid architecture and supports 55 languages. It was pre-trained on 2.8 million hours of data and has 2.88 billion parameters. The API has a boolean \texttt{remove\_disfluencies} option for handling filler words.

\paragraph{Whisper -- \texttt{large-v3}.} (\url{https://openai.com/index/whisper/}). Whisper is end-to-end transformer based ASR system released by OpenAI in 2022 \citep{radford_robust_2022} which received wide-spread attention for being open source. Whisper was trained on 680,000 hours of ``weakly'' supervised speech data (i.e. audio pulled from the internet with transcriptions that vary in quality), 117,000 hours of which was multilingual, covering 96 different languages. Studies have found that Whisper handles accented speech well but accuracy varies depending on the L1 background \citep{graham_evaluating_2024}. Many different platforms offer Whisper APIs. This study used the API offered by Replicate (\url{https://replicate.com/openai/whisper}), a cloud-based machine learning platform. There are different sized Whisper models with different numbers of parameters and computational requirements; smaller models run faster while larger models are more accurate. This study used \texttt{large-v3}, the largest and most accurate model with 1.54 billion parameters. 

Whisper does not have an option to include or exclude filler words like ``um'' and ``uh'', but it is possible to provide an initial prompt that contains additional context. The initial prompt can be used to supply obscure or technical words that are likely to appear in the audio. Including an initial prompt can make Whisper more likely to include disfluencies in transcriptions; however, it often causes Whisper to hallucinate and include words or expressions that are not in the audio. How the initial prompt was adjusted for this study to improve disfluency handling and reduce hallucinations is outlined in the discussion section.

\begin{table}
 \caption{ASR System Summary}
  \centering
  \small  
  \begin{tabularx}{\textwidth}{lXXXXX}
    \toprule
    Component & \multicolumn{5}{c}{ASR System} \tabularnewline
    \midrule
    & \centering AssemblyAI \texttt{Universal-2} & \centering Deepgram \texttt{Nova-2} & \centering RevAI \texttt{V2} & \centering Speechmatics \texttt{Ursa 2} & \centering Whisper \texttt{large-v3} \tabularnewline
    \midrule
    Approach & \centering RNN-T & \centering Transformer & \centering Conformer/CTC & \centering Hybrid & \centering Transformer \tabularnewline
    \midrule
    Parameters & \centering 600M & \centering Not disclosed & \centering 219.3M (encoder) & \centering 2.88B & \centering 1.54B \tabularnewline
    \midrule
    Supervised Training Data & \centering 300K hours supervised & \centering 47B tokens & \centering >3M hours & \centering Not disclosed & \centering 680K hours weakly supervised \tabularnewline
    \midrule
    Unsupervised Training Data & \centering 12.5M hours unsupervised audio & \centering Not disclosed & \centering Not disclosed & \centering 2.8M hours & \centering None \tabularnewline
    \midrule
    Number of Languages & \centering 20 & \centering 37 & \centering 59 & \centering 55 & \centering 96 \tabularnewline
    \midrule
    Filler Words Option & \centering Boolean  & \centering Boolean  & \centering Boolean  & \centering Boolean  & \centering Via initial prompt \tabularnewline
    \bottomrule
  \end{tabularx}
  \label{tab:table1}
\end{table}

\subsection{Metrics}

\paragraph{ASR Accuracy.}\textit{MER} was calculated using the \texttt{jiwer} Python package \citep{vaessen_jiwer_2018} by comparing the ASR transcription to the original human transcription of each speech sample. MERs for each transcription were collected and used to calculate totals and means for each analysis.

\paragraph{Processing Time.}In Python, a timer was started at the beginning of each API call and stopped when the response was received. Times for each transcription were recorded and used to calculate total and mean processing times.

\paragraph{Disfluency Handling.} Spoken disfluencies were categorized as fillers, repetitions, and revisions; empty pauses were not transcribed and thus not considered. Fillers were operationalized as the two tokens ``um'' and ``uh''. All instances of ``um'' and ``uh'' were counted in both the ground truth and ASR transcriptions, including cases where ASR systems detected fillers not present in the ground truth. Repetitions were operationalized as the immediate and exact repetition of one or two words such as ``the the'' and ``on the on the'' Only repetitions found in the ground truth and retained in the ASR transcriptions were evaluated. 

Filler detection rates and repetition retention rates were calculated differently to reflect their distinct evaluation approaches. Filler detection rates were calculated by dividing the number of fillers found in ASR transcriptions by the number identified in the ground truth (157 fillers across all recordings), which could exceed 1.0 when systems detected fillers not present in the ground truth. Repetition retention rates were calculated by dividing the number of ground truth repetitions preserved in ASR transcriptions by the total number of repetitions in the ground truth (40 repetitions). For example, if an ASR system identified 112 fillers compared to 157 in the ground truth, its filler detection rate would be 0.713, while if it preserved 31 out of 40 repetitions, its repetition retention rate would be 0.775.

Revisions were more complicated, and were operationalized as instances where the speaker reformulates an utterance by making a new attempt to express the same idea. Revision boundaries were typically set from the first word that appears in both attempts to the last word of the reformulation and includes everything in between such as in ``of um a downtown of a city''(speaker ABA). In this example, the revision begins at ``of'' because it appears in both attempts and ends with ``city'' as the revision of ``downtown''. However, in some cases, the shared word appeared at the end of the attempts rather than the beginning such as in ``the real their real'', in which case the revision spans from the first word of the initial attempt to the last word of the final attempt. False starts, where a speaker began a word or phrase but stopped mid-way and reformulated, were also considered revisions, such as in ``from s from somewhere'' (speaker EBVS) where the speaker stopped after making an [\textit{s}] sound and restarted the phrase. Some revisions included multiple attempts, such as in ``he found that he took uh he found'' (speaker ABA) or ``go to their go uh go to their destinations oh go to their home'' (speaker BWC). The ASR transcriptions of each revision were extracted and compared to the ground truth to calculate MER. This MER was converted to a revision accuracy score ($accuracy = 1 - MER$) between zero and one. A list of the ground truth and each ASR systems’ revision strings for can be found as supplementary materials on the OSF page, and the transcripts of speaker NCC with all disfluencies labeled can be seen in Appendix C.

When different categories of disfluencies overlapped, they were counted separately. For example, ``uh uh'' was counted as two filler words and one repetition. When reviewing the ground truth transcriptions, there were several ambiguous instances such as a schwa [\textit{\textschwa}] sound that could have been either ``a'' or ``uh''. Because these were indistinguishable, they were left as they were in the original transcriptions.

\section{Results}

\subsection{RQ1a -- Read Speech}
Research question 1a examined the accuracy and processing time of the ASR systems on recordings of read speech. A summary of mean MER by L1, gender, and ASR model can be found in~\ref{app:A}.

\subsubsection{Accuracy}
\begin{figure}
  \centering
  \includegraphics[width=1\textwidth]{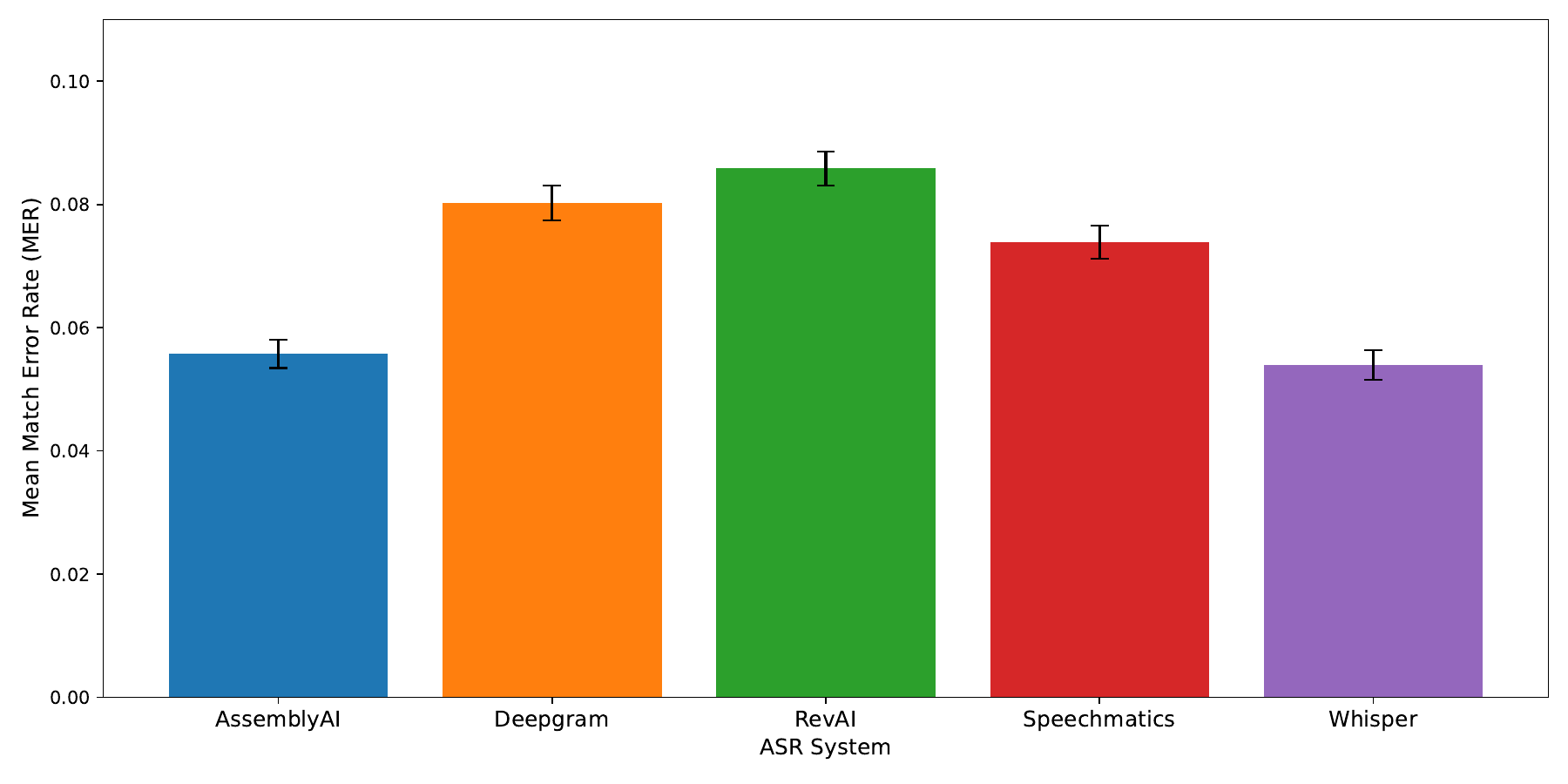}
  \caption{Read Speech: Mean \textit{MER} by ASR System\\
  \small\textit{Note.} Error bars represent standard error.}
  \label{fig:figure1}
\end{figure}

First, the mean \textit{MER} across all 2400 non-native English speech samples for each ASR was calculated to establish overall accuracy, seen in Table~\ref{tab:table2}. The majority of the read speech transcriptions had \textit{MER} scores of zero, and those with errors had a wide range. Because the \textit{MER} data was highly skewed, a Friedman test was conducted with the \texttt{R} programming language \citep{r_core_team_r_2024} in \texttt{RStudio} \citep{rstudio_team_rstudio_2024} to compare the effect of ASR system on \textit{MER}, finding a statistically significant difference with a small effect size ($\chi^2 = 471.93$, $df = 4$, $p < .001$, Kendall's $W = 0.039$). Post-hoc comparisons using Nemenyi's test indicated significant differences between ASR systems, and effect sizes ($r$) between systems were calculated based on mean rank differences, seen in Table~\ref{tab:table2}. Whisper ($M = 0.054$, $SD = 0.117$) and AssemblyAI ($M = 0.056$, $SD = 0.112$) performed significantly better than other models, with no significant difference between them ($p = .904$). RevAI showed the highest error rate ($M = 0.086$, $SD = 0.136$), performing significantly worse than both Whisper ($p < .001$) and AssemblyAI ($p < .001$). Deepgram ($M = 0.080$, $SD = 0.138$) and Speechmatics ($M = 0.074$, $SD = 0.132$) showed intermediate performance levels with no significant difference between them ($p = .879$). While the differences between models were statistically significant, the effect sizes based on mean rank differences were generally small, indicating limited practical significance. Figure~\ref{fig:figure1} shows a bar plot (chosen due to the highly skewed distribution) of the mean \textit{MER} for each ASR system with standard error bars.

\begin{table}
 \caption{Read Speech: ASR System Performance Comparison and Mean Rank Effect Size}
  \centering
  \begin{tabularx}{\textwidth}{lXccccc}
    \toprule
    & & \multicolumn{5}{c}{Effect size \textit{r}} \tabularnewline
    \cmidrule(r){3-7}
    ASR System & Mean \textit{MER} (\textit{SD}) & AssemblyAI & Deepgram & RevAI & Speechmatics & Whisper \tabularnewline
    \midrule
    AssemblyAI & 0.056 (0.112) & --- & 0.137* & 0.182* & 0.118* & 0.018 \tabularnewline
    Deepgram & 0.080 (0.138) & 0.137* & --- & 0.045 & 0.019 & 0.155* \tabularnewline
    RevAI & 0.086 (0.136) & 0.182* & 0.045 & --- & 0.064* & 0.200* \tabularnewline
    Speechmatics & 0.074 (0.132) & 0.118* & 0.019 & 0.064* & --- & 0.136* \tabularnewline
    Whisper & 0.054 (0.117) & 0.018 & 0.155* & 0.200* & 0.136* & --- \tabularnewline
    \bottomrule
    \multicolumn{7}{l}{\small Note. * Nemenyi's test \textit{p} < 0.05.}
  \end{tabularx}
  \label{tab:table2}
\end{table}

\subsubsection{L1 and Gender Effects}
\begin{figure}
  \centering
  \includegraphics[width=1\textwidth]{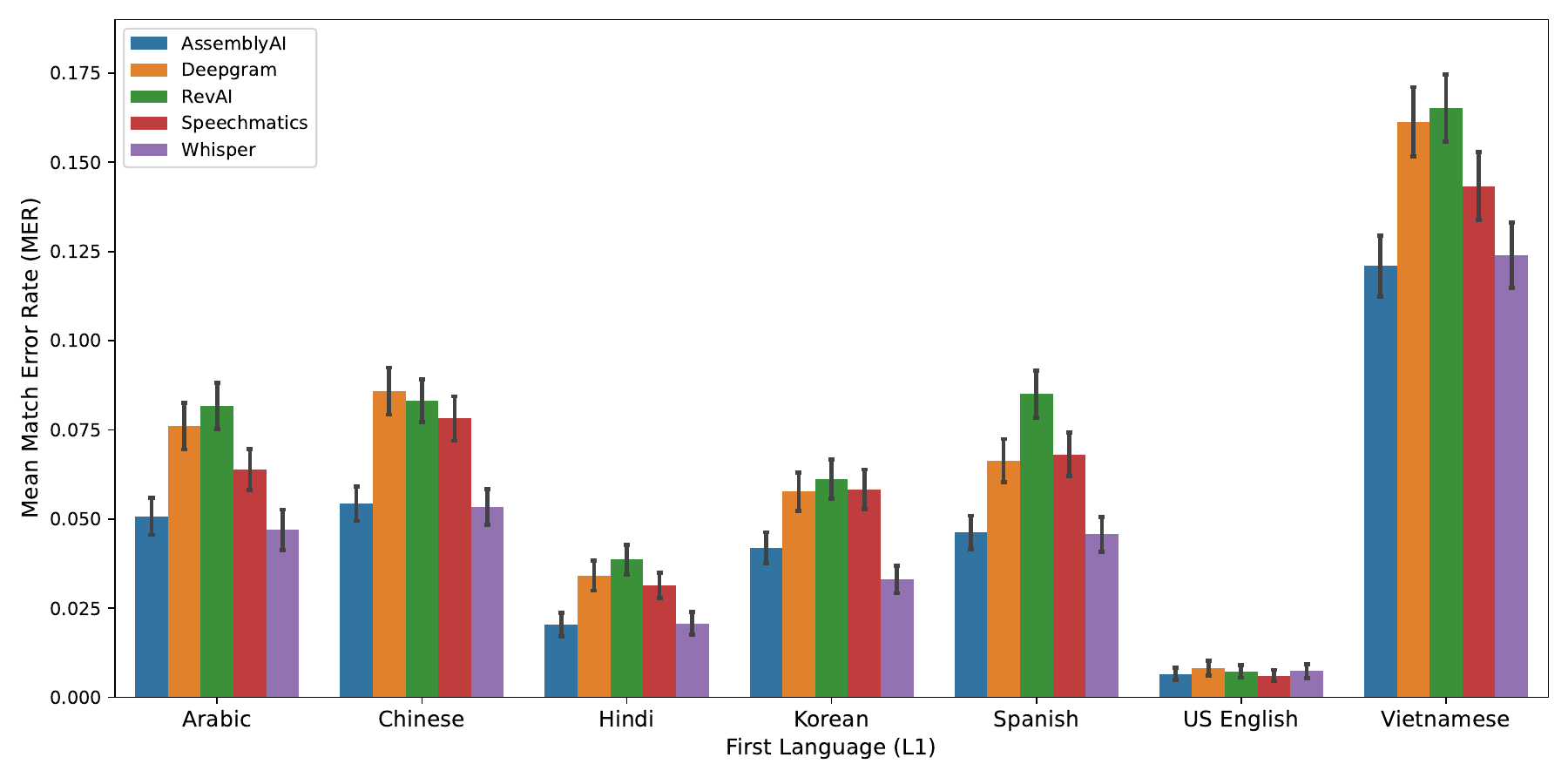}
  \caption{Read Speech: Mean \textit{MER} by L1 and ASR System\\
  \small\textit{Note.} Error bars represent standard error.}
  \label{fig:figure2}
\end{figure}

\begin{figure}
  \centering
  \includegraphics[width=1\textwidth]{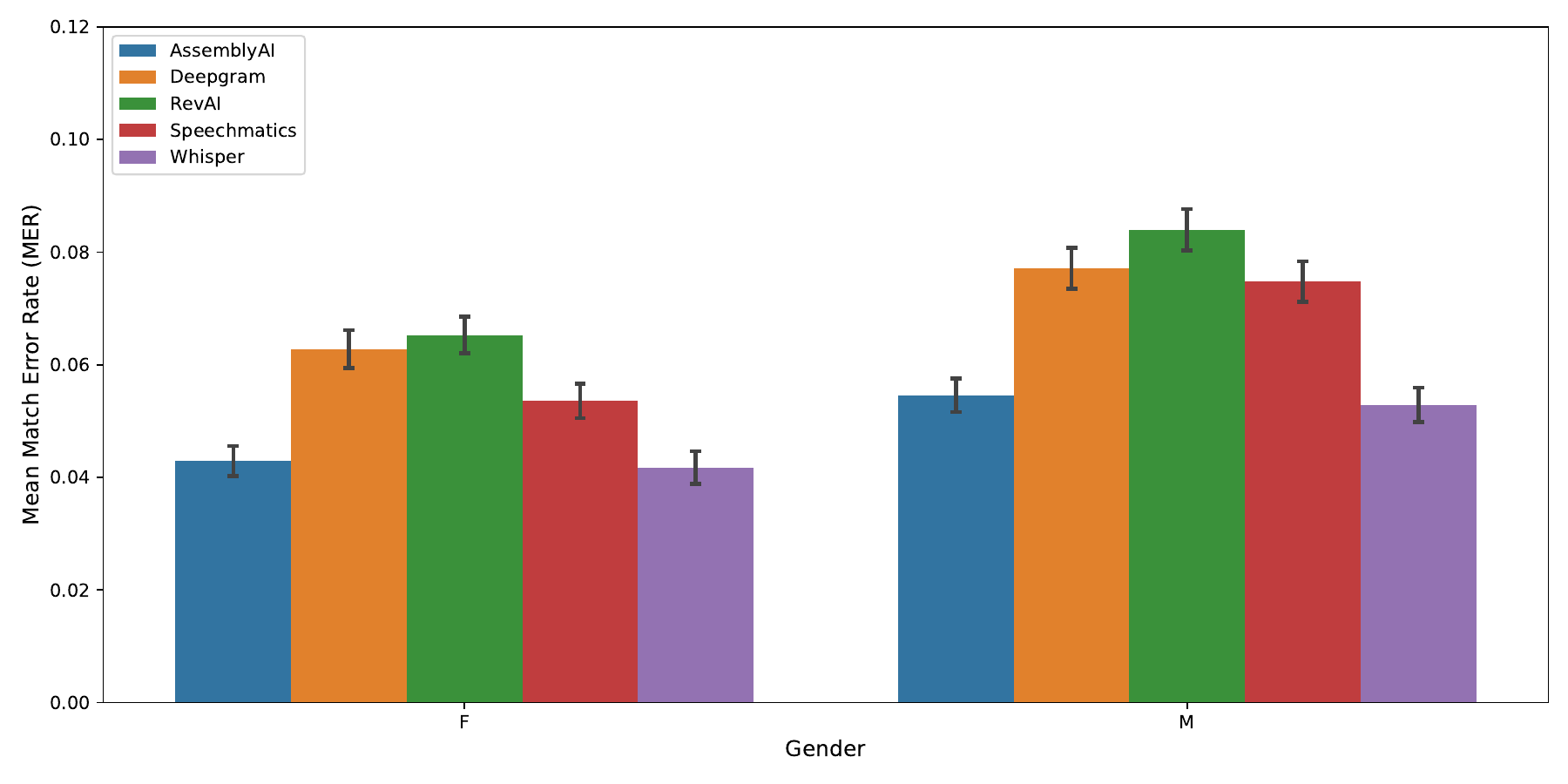}
  \caption{Read Speech: Mean \textit{MER} by Gender and ASR System\\
  \small\textit{Note.} Error bars represent standard error.}
  \label{fig:figure3}
\end{figure}

To examine potential differences in ASR performance across speaker demographics, mean \textit{MER} was calculated for each L1 background and gender. For read speech only, four US English speakers (two male, two female) from the CMU\_ARCTIC database \citep{kominek_cmu_2004}, contributing an additional 400 speech samples, were included as a control group to provide a native speaker reference point. Figure~\ref{fig:figure2} shows the mean \textit{MER} by L1 background for each ASR system, which are also listed in Table~\ref{tab:table3}. The pattern of relative performance among ASR systems remained consistent across L1 groups, with Whisper and AssemblyAI showing generally lower error rates than other systems. However, the magnitude of errors varied substantially across L1 groups, with US English speakers showing the lowest error rates ($M = 0.007$, $SD = 0.036$) and Vietnamese speakers showing notably higher error rates ($M = 0.143$, $SD = 0.186$) across all systems. Figure~\ref{fig:figure3} shows the mean \textit{MER} by gender for each ASR system. The relative performance of ASR systems remained similar across gender groups, though gender differences varied considerably by L1 group, with the largest differences observed in Vietnamese (male $M = 0.181$, $SD = 0.194$; female $M = 0.105$, $SD = 0.170$) and Spanish speakers (male $M = 0.084$, $SD = 0.131$; female $M = 0.041$, $SD = 0.091$).

\begin{table}
 \caption{Read Speech: Mean \textit{MER} and \textit{SD} by L1 and ASR System}
  \centering
  \setlength{\tabcolsep}{4pt}  
  \begin{tabularx}{\textwidth}{lXXXXX}
    \toprule
    L1 & AssemblyAI (\textit{SD}) & Deepgram (\textit{SD}) & RevAI (\textit{SD}) & Speechmatics (\textit{SD}) & Whisper (\textit{SD}) \\
    \midrule
    Arabic & 0.051 (0.104) & 0.076 (0.131) & 0.082 (0.129) & 0.064 (0.114) & 0.047 (0.113) \\
    Chinese & 0.054 (0.096) & 0.086 (0.132) & 0.083 (0.120) & 0.078 (0.125) & 0.053 (0.099) \\
    Hindi & 0.020 (0.066) & 0.034 (0.085) & 0.039 (0.082) & 0.031 (0.071) & 0.021 (0.063) \\
    Korean & 0.042 (0.088) & 0.058 (0.107) & 0.061 (0.111) & 0.058 (0.112) & 0.033 (0.077) \\
    Spanish & 0.046 (0.094) & 0.066 (0.120) & 0.085 (0.132) & 0.068 (0.122) & 0.046 (0.098) \\
    US English & 0.007 (0.034) & 0.008 (0.041) & 0.007 (0.035) & 0.006 (0.031) & 0.007 (0.040) \\
    Vietnamese & 0.121 (0.171) & 0.161 (0.194) & 0.165 (0.187) & 0.143 (0.191) & 0.124 (0.183) \\
    \bottomrule
  \end{tabularx}
  \label{tab:table3}
\end{table}

A repeated-measures ANOVA was conducted in \texttt{RStudio} using the \texttt{ez} library \citep{lawrence_ez_2016} to investigate the impact of L1, gender, and ASR system on \textit{MER}, with ASR system as a within-subjects factor and L1 and gender as between-subjects factors: \textit{MER} $\sim$ ASR System $\times$ L1 $\times$ gender. The model found significant main effects for ASR system ($F(4, 56) = 64.37$, $p < .001$, $\mathit{ges} = 0.217$) and L1 ($F(6, 14) = 8.71$, $p < .001$, $\mathit{ges} = 0.778$), but no significant effect of gender ($F(1, 14) = 2.00$, $p = .179$, $\mathit{ges} = 0.118$). Mauchly's test indicated a violation of sphericity ($W = 0.13$, $p = .003$), but the ASR system effect remained significant with Greenhouse-Geisser correction ($\varepsilon = 0.51$, $p < .001$). A significant interaction was found between L1 and ASR system ($F(24, 56) = 3.24$, $p < .001$, $\mathit{ges} = 0.077$), while other interactions were not significant.

Analysis of L1 differences revealed a clear progression in ASR performance across speaker groups. US English speakers had significantly lower error rates than all non-native groups (all $p < .001$). Among non-native speakers, Vietnamese speakers had significantly higher error rates compared to all other L1 groups (all $p < .001$), with the largest difference observed between Vietnamese and Hindi speakers (estimate = 0.114, $p < .001$). Hindi speakers showed significantly lower error rates compared to all other non-native groups (all $p < .001$). Korean speakers differed significantly from both Hindi ($p < .001$) and Spanish ($p = .028$) speakers. No significant differences were found between Arabic, Chinese, and Spanish speakers (all $p > .22$). A significant interaction between L1 and ASR system ($F(24, 56) = 3.24$, $p < .001$, $\mathit{ges} = 0.077$) indicated that the pattern of ASR performance varied across L1 groups, with differences between ASR systems being more pronounced for speakers with higher overall error rates.

Although gender differences were not statistically significant, male speakers consistently had higher error rates than female speakers across all ASR systems, with female speakers having a mean \textit{MER} of 0.053 ($SD = 0.115$) compared to male speakers' mean \textit{MER} of 0.069 ($SD = 0.127$). The magnitude of this gender difference varied across systems: Speechmatics showed the largest difference (0.025 higher \textit{MER} for males), followed by RevAI (0.022), while AssemblyAI and Whisper showed the smallest differences (0.013 each). Gender differences also varied considerably by L1 group, with the largest differences observed in Vietnamese (male $M = 0.181$, $SD = 0.194$; female $M = 0.105$, $SD = 0.170$) and Spanish speakers (male $M = 0.084$, $SD = 0.131$; female $M = 0.041$, $SD = 0.091$). Despite these descriptive variations, the gender interaction was not significant ($F(4, 56) = 2.04$, $p = .103$, $\mathit{ges} = 0.01$), suggesting that the pattern of gender differences did not meaningfully vary by ASR system.

\subsubsection{Processing Time} 
While processing time may vary based on server conditions, analysis of the API calls revealed some notable differences between ASR systems. The API calls were timed in \texttt{Python} and mean times were calculated for the non-native read speech samples (US English speech samples were not included in this analysis). Mean processing times are shown in Table ~\ref{tab:table4} below. Figure ~\ref{fig:figure4} compares accuracy and time for all measurements as well as the means for each ASR system. To examine the relationship between accuracy and processing time, a Spearman's rank correlation was conducted between \textit{MER} and processing time across all systems, finding no significant correlation ($\rho = -0.002$, $p = .855$). This suggests that longer processing time does not necessarily result in better accuracy.

An efficiency metric for each transcription was then calculated by dividing accuracy (1 - \textit{MER}) by processing time, yielding a measure of accuracy per second, also seen in Table ~\ref{tab:table4}. A one-way ANOVA found significant differences in efficiency between ASR systems ($F(4, 11995) = 15966$, $p < .001$). Deepgram showed the highest efficiency ($M = 0.509$, $SD = 0.103$), with the fastest processing time despite relatively moderate accuracy. RevAI showed the lowest efficiency ($M = 0.111$, $SD = 0.021$), with both slower processing and higher error rates. The remaining systems --- Whisper ($M = 0.226$, $SD = 0.053$), AssemblyAI ($M = 0.208$, $SD = 0.047$), and Speechmatics ($M = 0.198$, $SD = 0.033$) --- showed intermediate efficiency levels.

\begin{table}
 \caption{Read Speech: Mean Processing Time in Seconds and Efficiency by ASR system}
  \centering
  \setlength{\tabcolsep}{4pt}
  \begin{tabularx}{\textwidth}{lXX}
    \toprule
    Model & Processing Time $M$ ($SD$) & Efficiency $M$ ($SD$) \\
    \midrule
    AssemblyAI & 5.045 (3.806) & 0.208 (0.047) \\
    Deepgram & 1.843 (0.289) & 0.509 (0.103) \\
    RevAI & 8.649 (3.746) & 0.111 (0.021) \\
    Speechmatics & 4.705 (0.405) & 0.198 (0.033) \\
    Whisper & 4.426 (2.355) & 0.226 (0.053) \\
    \bottomrule
    \multicolumn{3}{l}{\small\textit{Note.} $n = 2400$ transcriptions per model}
  \end{tabularx}
  \label{tab:table4}
\end{table}

\begin{figure}
  \centering
  \includegraphics[width=1\textwidth]{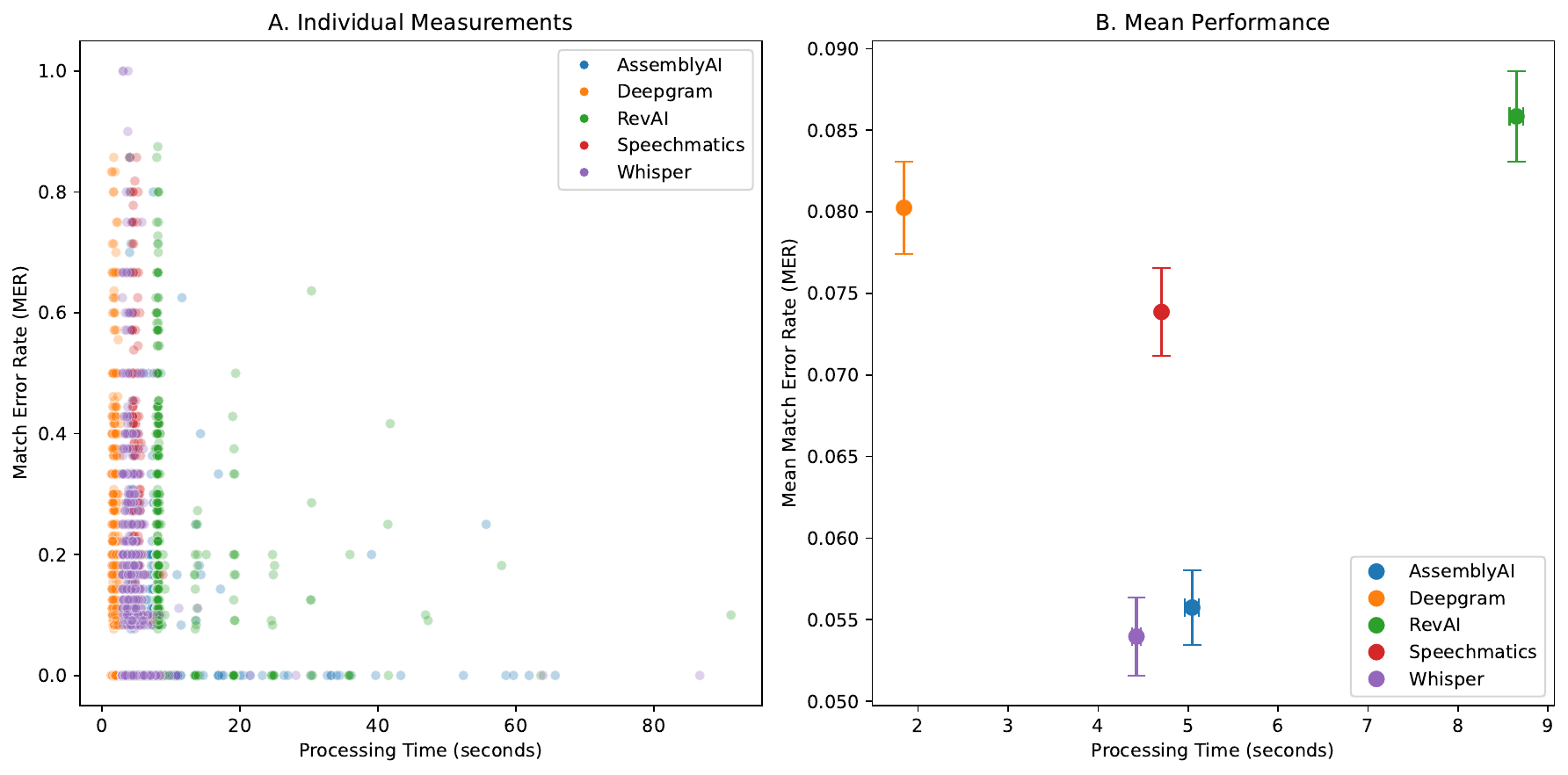}
  \caption{Read Speech: Accuracy vs. Processing Time by ASR System\\[1ex]
  \protect\small\textit{Note.} Error bars represent standard error.}
  \label{fig:figure4}
\end{figure}

Post-hoc comparisons using Tukey's HSD test indicated that all pairwise differences were significant ($p < .001$). Effect size calculations found very large differences between Deepgram and other systems (Cohen's $d$ ranging from 3.46 to 5.34), while differences among AssemblyAI, Speechmatics, and Whisper were more modest (Cohen's $d$ ranging from 0.25 to 0.63). These results suggest that while Whisper and AssemblyAI achieved the highest accuracy overall, Deepgram's processing speed made it the most efficient system when considering both speed and accuracy together.

\subsection{RQ1b -- Spontaneous Speech}
Research question 1b looked at the accuracy and processing time of the ASR systems on recordings of spontaneous speech. ASR transcriptions of spontaneous speech commonly have higher error rates than read speech due to its naturalistic nature and the increased presence of disfluencies. While the read speech data from L2-ARCTIC consisted of 2,400 short recordings (from 0.8 to 7.9 seconds, averaging 2.98 seconds) of identical sentences for each speaker, the spontaneous narrative recordings were considerably longer and more variable in length, ranging from 27.5 to 235 seconds (averaging 71.8 seconds) with a total duration of 1,566 seconds (26 minutes). 

\subsubsection{Accuracy and Disfluency Condition} 
\begin{figure}
  \centering
  \includegraphics[width=1\textwidth]{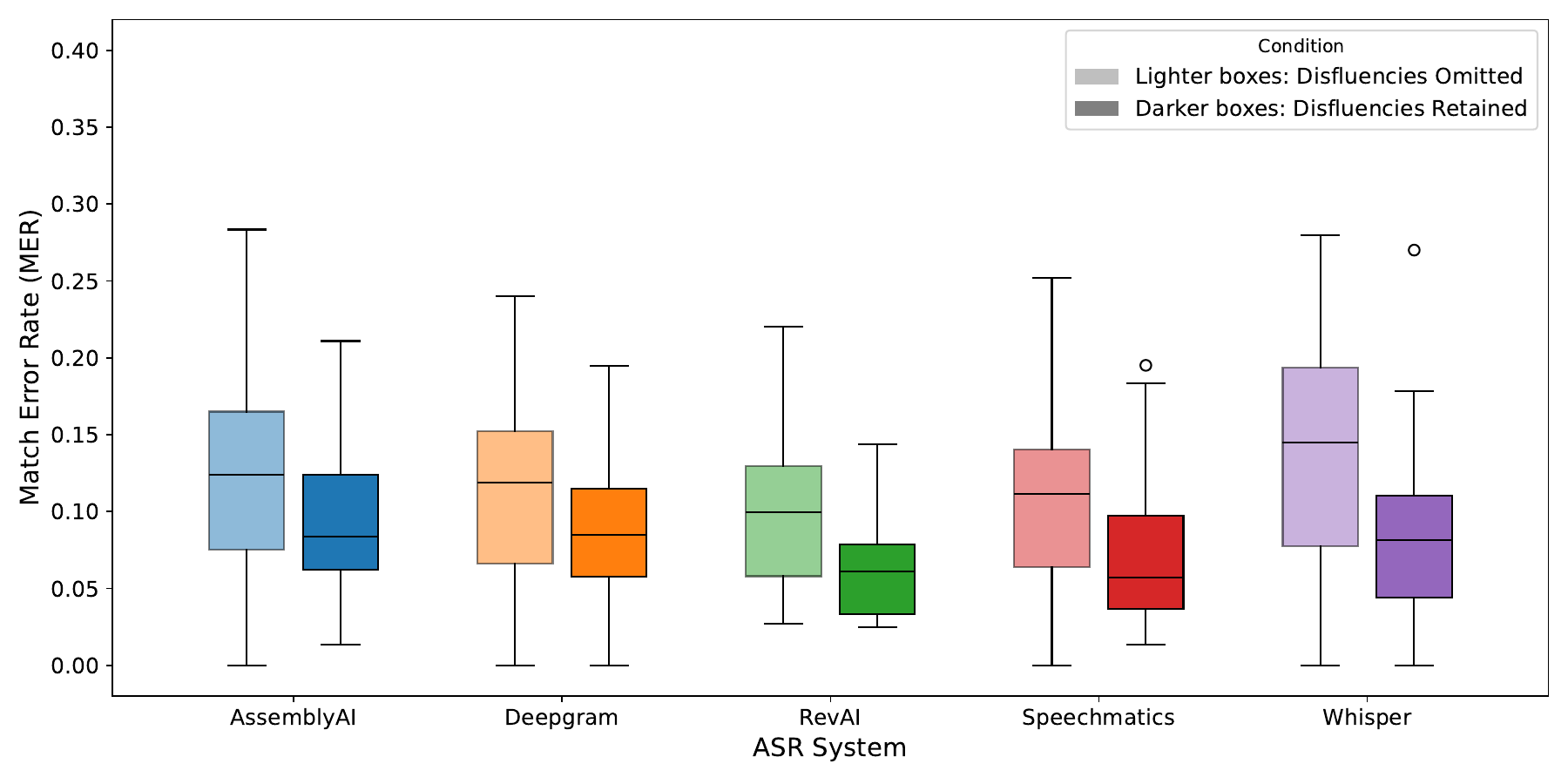}
  \caption{Spontaneous Speech: \textit{MER} Distribution by ASR System and Disfluency Condition\\}
  \label{fig:figure5}
\end{figure}

First, the recordings were transcribed by each ASR system under both disfluency conditions (retaining and omitting filler words) to see how performance was affected. The \textit{MER}s for each transcription by the five ASR systems under both conditions can be found in~\ref{app:B}. Figure~\ref{fig:figure5} shows \textit{MER} distribution by ASR system and disfluency condition.

A repeated-measures ANOVA was conducted to investigate the impact of ASR system and disfluency condition on \textit{MER}, with both factors as within-subjects factors: \textit{MER} $\sim$ ASR system $\times$ condition. Significant main effects were found for both ASR system ($F(4,84) = 5.83$, $p < .001$, $\mathit{ges} = 0.039$) and disfluency condition ($F(1,21) = 28.76$, $p < .001$, $\mathit{ges} = 0.092$). No significant interaction was found between ASR system and disfluency condition ($F(4,84) = 2.33$, $p = .062$, $\mathit{ges} = 0.006$). Given violations of sphericity in the interaction term (Mauchly's $W = 0.004$, $p < .001$), Greenhouse-Geisser corrections were applied but did not change the pattern of significance. Post-hoc pairwise comparisons with Bonferroni correction revealed significant differences between RevAI and Whisper (estimate = 0.033, $p < .001$) and between AssemblyAI and RevAI (estimate = 0.026, $p = .015$). The difference between Speechmatics and Whisper approached significance (estimate = 0.022, $p = .062$).

The effect of the disfluency condition was examined using paired $t$-tests for each ASR system, with all systems showing significantly lower mean \textit{MER} when retaining disfluencies compared to omitting them. As seen in Table~\ref{tab:table5}, Whisper showed the largest absolute difference in \textit{MER} between conditions (0.051), followed by RevAI (0.040), Speechmatics (0.037), Deepgram (0.028), and AssemblyAI (0.026). In terms of statistical strength, RevAI showed the most robust effect ($t = 4.84$, $p < .0001$), followed by Speechmatics ($t = 4.19$, $p < .001$), Deepgram ($t = 4.16$, $p < .001$), Whisper ($t = 3.95$, $p < .001$), and AssemblyAI ($t = 3.82$, $p < .001$). Standard deviations were generally larger in the disfluencies omitted condition, suggesting more variable performance when ASR systems omitted disfluencies from the transcription. Because the ASR transcriptions that retained disfluencies showed better performance, they were used for the rest of the analyses.

\begin{table}
 \caption{ASR System Mean \textit{MER} and Difference by Disfluency Condition}
  \centering
  \small  
  \setlength{\tabcolsep}{3pt}  
  \begin{tabular}{lccccccc}  
    \toprule
    & \multicolumn{2}{c}{Disfluencies Omitted} & \multicolumn{2}{c}{Disfluencies Retained} & & & \\
    \cmidrule(r){2-3} \cmidrule(r){4-5}
    ASR System & $M$ & $SD$ & $M$ & $SD$ & $M$ Difference & $t$-value & $p$-value \\
    \midrule
    AssemblyAI & 0.122 & 0.068 & 0.096 & 0.050 & 0.026 & 3.820 & $<$0.001 \\
    Deepgram & 0.114 & 0.059 & 0.085 & 0.046 & 0.028 & 4.160 & $<$0.001 \\
    RevAI & 0.103 & 0.055 & 0.063 & 0.034 & 0.040 & 4.840 & $<$0.0001 \\
    Speechmatics & 0.112 & 0.067 & 0.075 & 0.054 & 0.037 & 4.190 & $<$0.001 \\
    Whisper & 0.142 & 0.079 & 0.090 & 0.061 & 0.051 & 3.950 & $<$0.001 \\
    \bottomrule
  \end{tabular}
  \label{tab:table5}
\end{table}

\subsubsection{L1 and Gender Effects} 
Due to two speakers (an L1 Arabic female and L1 Hindi male) not completing the narrative recording task, the groups were slightly unbalanced. Type III sums of squares were used in the analysis to account for this imbalance. A repeated measures ANOVA was conducted to examine the impact of L1, gender, and ASR system on \textit{MER}, with ASR system as a within-subjects factor and L1 and gender as between-subjects factors: \textit{MER} $\sim$ ASR system $\times$ L1 $\times$ gender. The analysis revealed no significant main effects of L1 ($F(5,10) = 2.32$, $p = .121$, $\mathit{ges} = 0.395$) or gender ($F(1,10) = 0.11$, $p = .749$, $\mathit{ges} = 0.006$). The main effect of ASR system was initially significant ($F(4,40) = 2.64$, $p = .048$, $\mathit{ges} = 0.103$), but after applying Greenhouse-Geisser correction for sphericity violation (Mauchly's $W = 0.100$, $p = .024$), it became non-significant ($p = .097$). No significant interactions were found between these factors (all $p > .20$).

\begin{figure}
  \centering
  \includegraphics[width=1\textwidth]{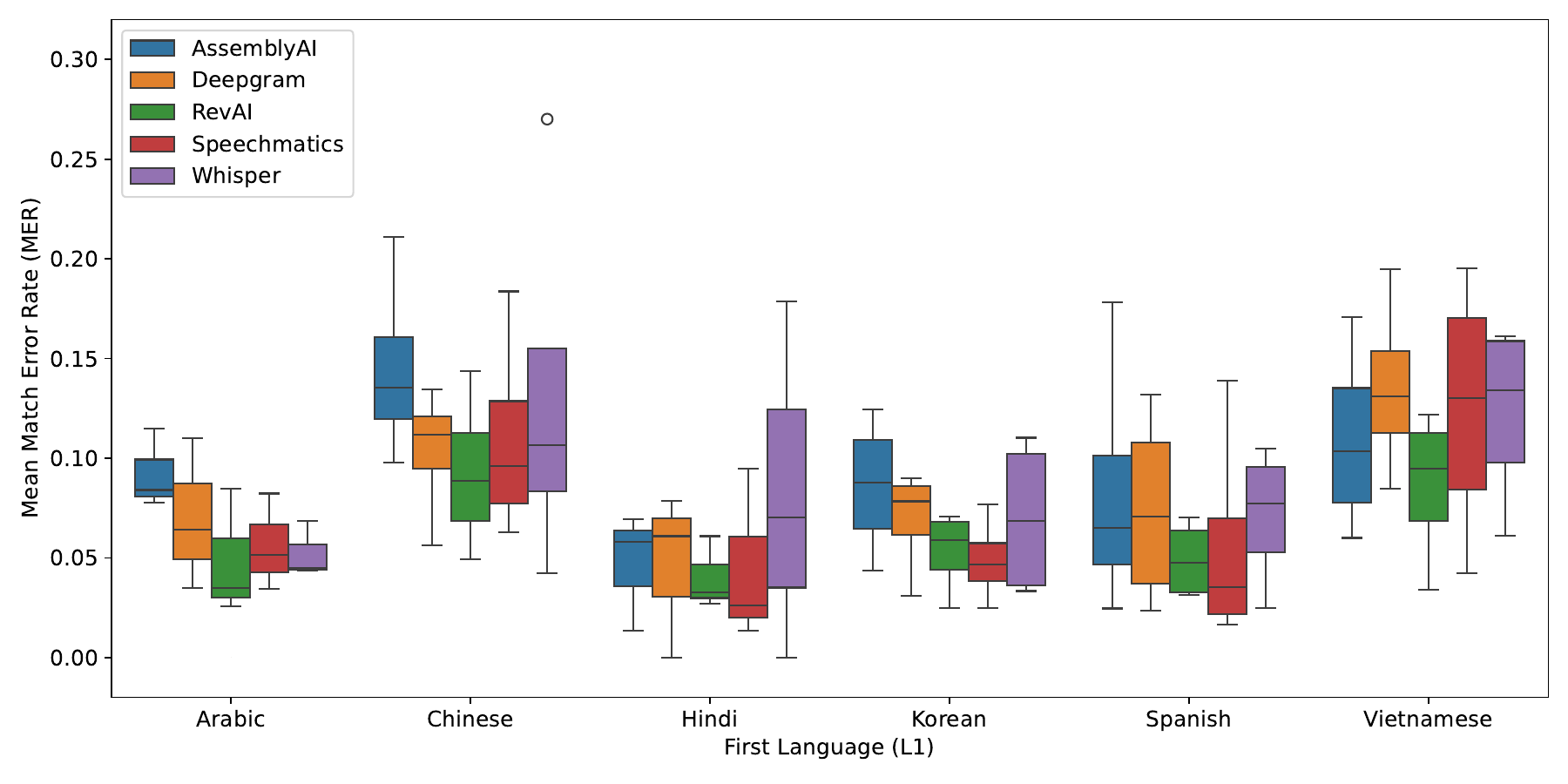}
  \caption{Spontaneous Speech: \textit{MER} Distribution by L1 and ASR System\\}
  \label{fig:figure6}
\end{figure}

As shown in Figure~\ref{fig:figure6}, while not reaching statistical significance, the distribution of \textit{MER} scores varied across L1 backgrounds. Speakers with Hindi L1 showed consistently low mean \textit{MER} scores (0.040--0.083) and minimal variability. In contrast, those with Chinese (\textit{means}: 0.093--0.145) and Vietnamese (\textit{means}: 0.086--0.135) L1 backgrounds showed notably higher scores and greater variability across ASR systems. Speakers with Arabic (\textit{means}: 0.048--0.092), Korean (\textit{means}: 0.053--0.086), and Spanish (\textit{means}: 0.049--0.083) L1 backgrounds showed intermediate distributions of scores.

\begin{figure}
  \centering
  \includegraphics[width=1\textwidth]{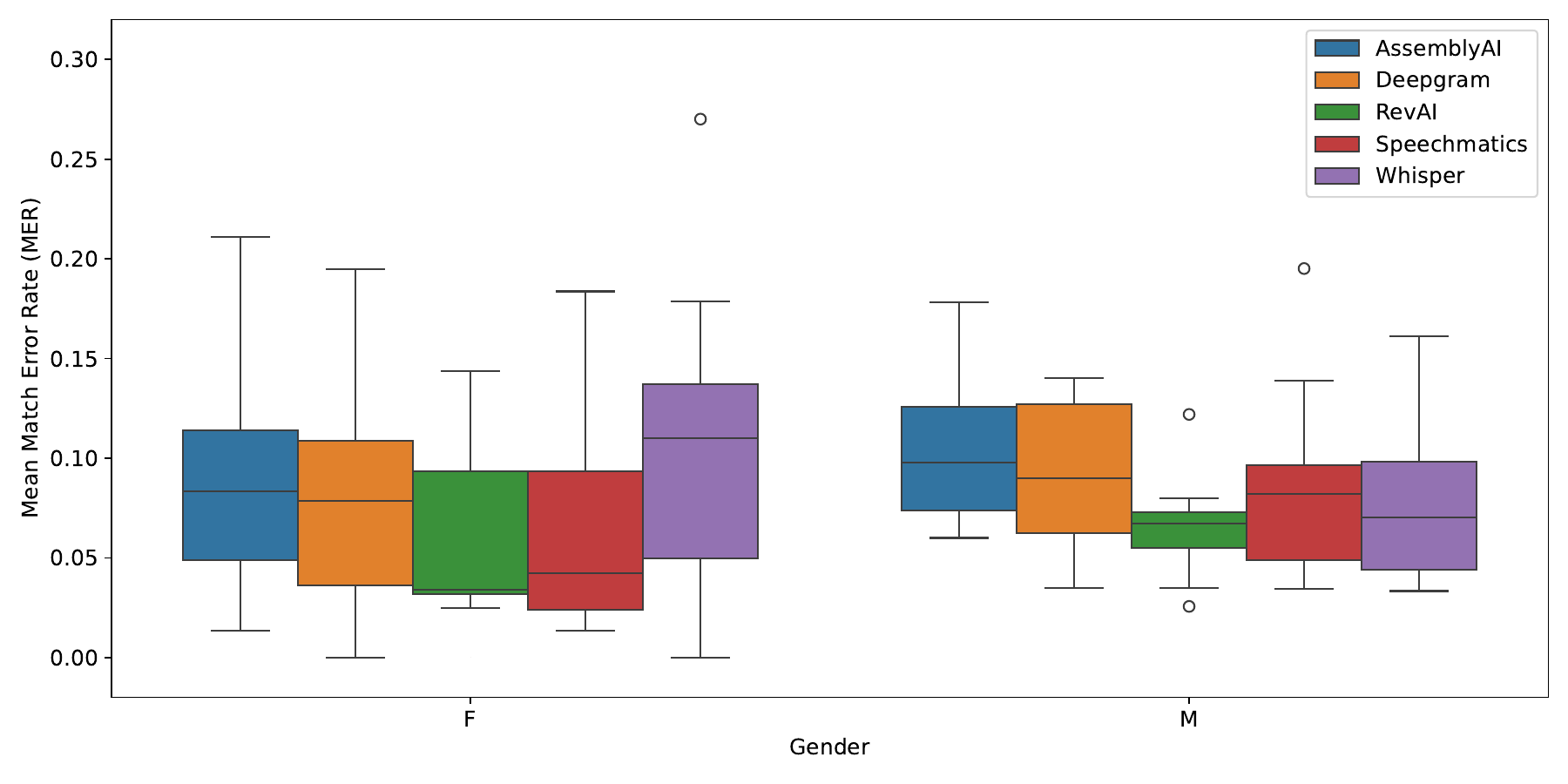}
  \caption{Spontaneous Speech: \textit{MER} Distribution by Gender and ASR System\\}
  \label{fig:figure7}
\end{figure}

Male and female speakers showed similar patterns of \textit{MER} distributions across ASR systems, seen in Figure~\ref{fig:figure7}, consistent with the lack of significant gender effect or interactions. While female speakers showed slightly more variability with Whisper ($M = 0.103$, $SD = 0.078$), and male speakers showed more consistent performance across systems (\textit{means} ranging from 0.065 for RevAI to 0.106 for AssemblyAI), these differences were not statistically significant. This pattern aligns with the read speech findings, where gender similarly showed no significant effect on ASR performance.

\subsubsection{Processing Time} 
\begin{figure}
  \centering
  \includegraphics[width=1\textwidth]{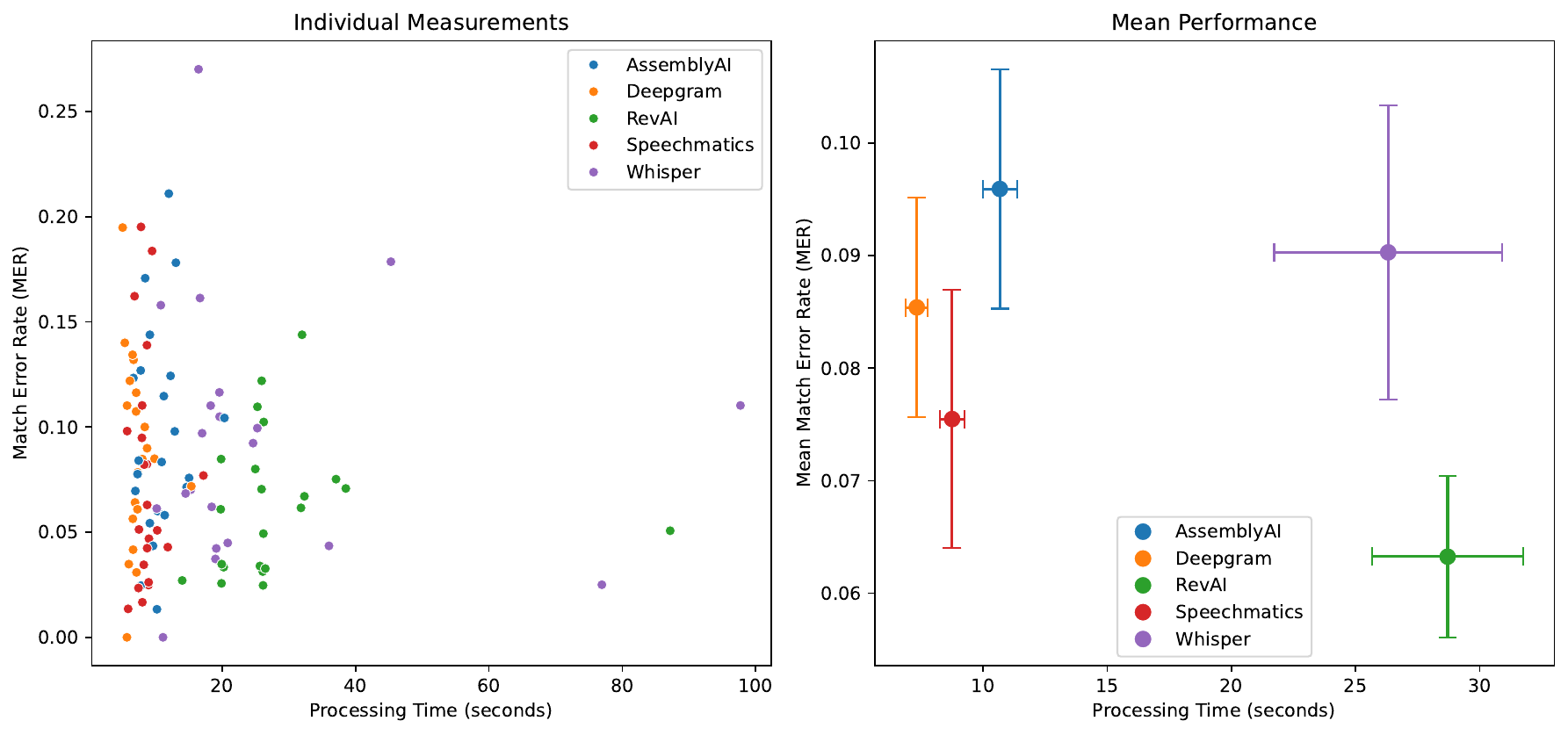}
  \captionsetup{justification=centering}
  \caption{Spontaneous Speech: Processing Time Distribution by ASR System\\[1ex]
  \protect\small\textit{Note.} Error bars in the right panel represent standard error.}
  \label{fig:figure8}
\end{figure}

The spontaneous narrative recordings varied substantially in length (ranging from 27.5 to 235 seconds, $M = 71.8$ seconds), unlike the uniform short recordings in the read speech task ($M = 2.98$ seconds). This wide range in recording duration naturally affected processing times, with longer files requiring more processing time from all ASR systems. As seen in Figure~\ref{fig:figure8}, processing times varied considerably across systems, with RevAI ($M = 28.71$, $SD = 14.31$) and Whisper ($M = 26.31$, $SD = 21.54$) showing particularly high processing times and variation, while Deepgram ($M = 7.34$, $SD = 2.12$), Speechmatics ($M = 8.76$, $SD = 2.30$), and AssemblyAI ($M = 10.68$, $SD = 3.26$) maintained more consistent and faster processing times.

To examine the relationship between accuracy and processing time, a Spearman's rank correlation was conducted between \textit{MER} and processing time across all systems, finding no significant correlation ($\rho = -0.087$, $p = .368$). This suggests that longer processing time does not necessarily result in better accuracy, similar to the pattern found in read speech.

An efficiency metric for each transcription was calculated by dividing accuracy (1 - \textit{MER}) by processing time, yielding a measure of accuracy per second. A one-way ANOVA found significant differences in efficiency between ASR systems ($F(4,105) = 73.19$, $p < .001$). Deepgram showed the highest efficiency ($M = 0.131$, $SD = 0.025$), maintaining fast processing times with moderate accuracy. RevAI showed the lowest efficiency ($M = 0.037$, $SD = 0.012$), due to longer processing times. The remaining systems showed varying levels of efficiency: Speechmatics ($M = 0.111$, $SD = 0.024$), AssemblyAI ($M = 0.092$, $SD = 0.026$), and Whisper ($M = 0.048$, $SD = 0.021$).

Post-hoc comparisons using Tukey's HSD test indicated that almost all pairwise differences were significant ($p < .05$), with only RevAI and Whisper showing no significant difference ($p = .517$). These results suggest that while RevAI achieved good accuracy, its slower processing time impacted its overall efficiency. Conversely, Deepgram's fast processing speed made it the most efficient system despite moderate accuracy scores.

\subsection{RQ2 -- Disfluency Handling}
\begin{figure}
  \centering
  \includegraphics[width=1\textwidth]{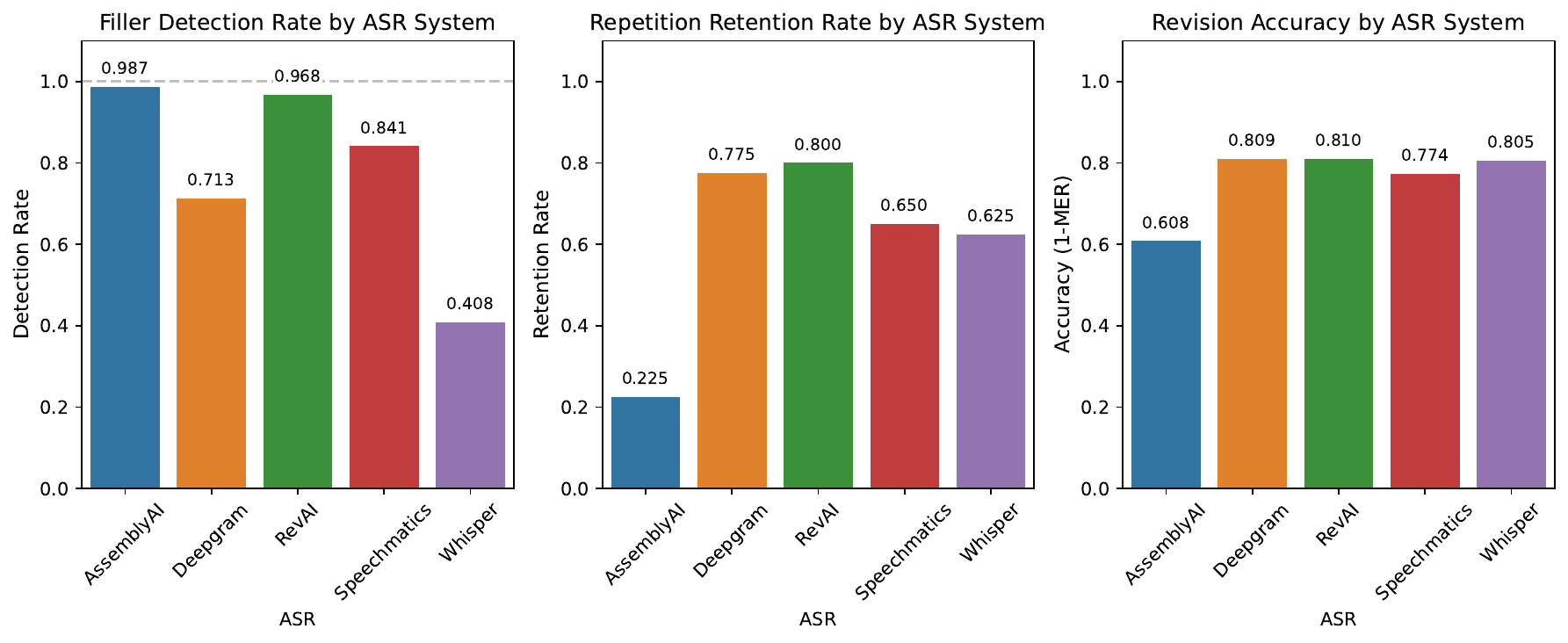}
  \captionsetup{justification=centering}
  \caption{Spontaneous Speech: Overall Disfluency Measurements by ASR System\\[1ex]
  \protect\small\textit{Note.} Filler and repetition bars represent overall rates (total detected/total in ground truth). Revision bars represent mean accuracy values (1-\textit{MER}) across all speakers.}
  \label{fig:figure9}
\end{figure}

To assess disfluency handling, only the ASR transcriptions with disfluencies retained were used. The three operationalized categories of disfluencies---fillers, repetitions, and revisions---were manually identified in the ground truth following the guidelines outlined in the methods. 157 fillers were identified in 20 of the ground truth transcriptions (two speakers did not use any fillers) ranging from 1 to 22 per speaker ($M = 7.85$). 40 repetitions were identified in 12 of the transcriptions (10 speakers did not have any repetitions) ranging from one to nine per speaker ($M = 3.34$). 54 revisions were identified in 18 of the transcriptions (four speakers did not have revisions) ranging from one to nine per speaker ($M = 3$). Descriptive statistics of filler and revision retention rates and revision \textit{MER}s can be seen in Table~\ref{tab:table6} and Figure~\ref{fig:figure9}.

\begin{table}
 \caption{ASR System Disfluency Measurements}
  \centering
  \setlength{\tabcolsep}{4pt}
  \begin{tabular}{lcccccc}
    \toprule
    & \multicolumn{2}{c}{Fillers (\textit{total} = 157)} & \multicolumn{2}{c}{Repetitions (\textit{total} = 40)} & \multicolumn{2}{c}{Revisions (\textit{total} = 54)} \\
    \cmidrule(r){2-3} \cmidrule(r){4-5} \cmidrule(r){6-7}
    ASR System & Count & Detection Rate & Count & Retention Rate & Mean \textit{MER} ($SD$) & Accuracy (1 - \textit{MER}) \\
    \midrule
    AssemblyAI & 155 & 0.987 & 9 & 0.225 & 0.392 (0.169) & 0.608 \\
    Deepgram & 112 & 0.713 & 31 & 0.775 & 0.191 (0.150) & 0.809 \\
    RevAI & 152 & 0.968 & 32 & 0.800 & 0.190 (0.172) & 0.810 \\
    Speechmatics & 132 & 0.841 & 26 & 0.650 & 0.226 (0.202) & 0.774 \\
    Whisper & 64 & 0.408 & 25 & 0.625 & 0.195 (0.170) & 0.805 \\
    \bottomrule
  \end{tabular}
  \label{tab:table6}
\end{table}

\subsubsection{Filler Detection} 
To compare filler detection across the five ASR systems, a chi-square test was conducted which found significant differences in performance ($\chi^2 = 208.28$, $df = 4$, $p < .001$). Post-hoc pairwise comparisons with Bonferroni correction showed that AssemblyAI and RevAI had the highest overall detection rates (0.987 and 0.968 respectively, with no significant difference between them, $p = 1.00$). However, analysis of individual transcriptions revealed instances where these systems detected fillers not present in the ground truth, with some recordings showing detection rates up to 2.5 times the actual number of fillers, as seen in the left panel of Figure~\ref{fig:figure10}. Speechmatics showed a moderate detection rate (0.841), significantly lower than both AssemblyAI ($p < .001$) and RevAI ($p < .01$). Deepgram showed a lower rate (0.713), significantly different from AssemblyAI and RevAI (both $p < .001$), though its difference from Speechmatics was borderline non-significant ($p = .099$). Whisper detected the fewest fillers (0.408), performing significantly worse than all other systems (all $p < .001$).

\subsubsection{Repetition Retention} 
Repetition retention distribution is shown in the center panel of Figure~\ref{fig:figure10}. A chi-square test found significant differences in repetition retention performance across the five ASR systems ($\chi^2 = 36.026$, $df = 4$, $p < .001$). Post-hoc pairwise comparisons with Bonferroni correction showed that RevAI and Deepgram had the highest retention rates (0.800 and 0.775 respectively, with no significant difference between them, $p = 1.00$). Speechmatics and Whisper showed moderate retention (0.650 and 0.625 respectively), and were not significantly different from RevAI and Deepgram (all $p = 1.00$). AssemblyAI showed substantially lower repetition retention (0.225), performing significantly worse than all other systems (all $p < .01$). Notably, while AssemblyAI excelled at filler retention, it struggled considerably with repetition retention.

\subsubsection{Revision Accuracy} 
\begin{figure}
  \centering
  \includegraphics[width=1\textwidth]{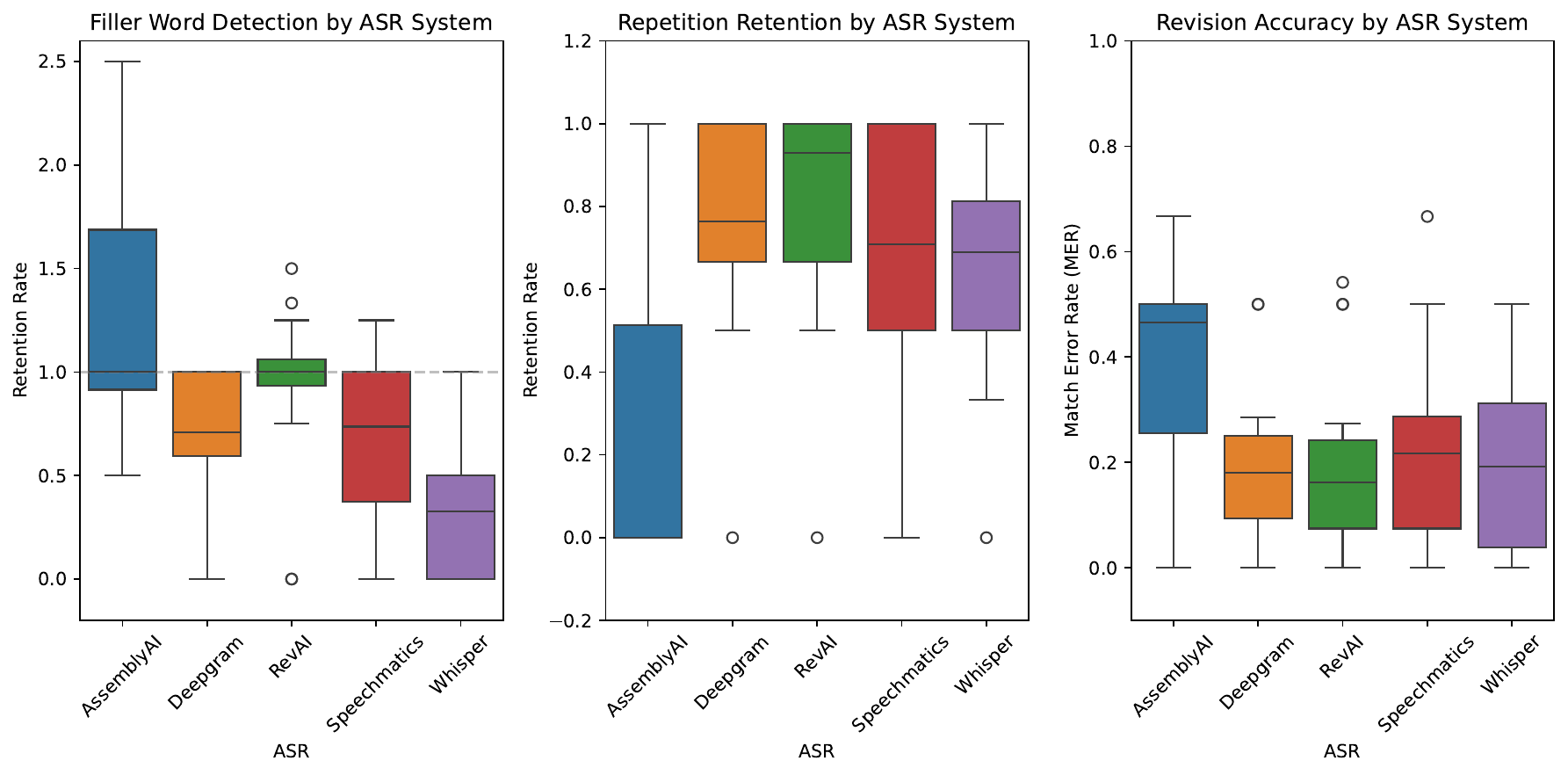}
  \caption{Spontaneous Speech: Disfluency Measurement Distribution by ASR System\\}
  \label{fig:figure10}
\end{figure}

Revision transcription accuracy varied significantly among the five ASR systems as demonstrated by a Friedman test ($\chi^2 = 24.156$, $df = 4$, $p < .001$) and visualized in the right panel of Figure~\ref{fig:figure10}. Post-hoc pairwise comparisons using the Nemenyi test showed that RevAI, Deepgram, and Whisper had the highest accuracy rates (0.810, 0.809, and 0.805 respectively), with no significant differences among them (all $p > .99$). Speechmatics showed slightly lower accuracy (0.774) but was not significantly different from the top performers (all $p > .96$). AssemblyAI showed substantially lower revision accuracy (0.608), performing significantly worse than all other systems (all $p < .05$). This pattern contrasts with filler retention, where AssemblyAI was among the top performers, demonstrating performance differences across disfluency types.

\subsubsection{Patterns Across Disfluency Types} 
Different ASR systems showed distinct patterns in handling various types of disfluencies. AssemblyAI showed the most variable performance, excelling at filler retention while performing poorly on repetitions and worst on revisions. RevAI demonstrated the most consistent performance across types, with near-perfect filler retention, high repetition retention, and the best revision accuracy. Whisper showed an opposite pattern to AssemblyAI for simpler disfluencies, performing poorly on fillers while achieving moderate accuracy on repetitions. Deepgram and Speechmatics generally showed similar patterns to each other across all three disfluency types.

\section{Discussion}
To the best of my knowledge, no peer-reviewed studies have assessed the performance of AssemblyAI, Deepgram, RevAI, or Speechmatics on non-native speech, but some studies have examined Whisper. The only study that reports \textit{MER} scores which can be directly compared is \citet{graham_evaluating_2024} which tested Whisper-large-v3 on English read speech from speakers from 14 different L1 backgrounds. They found a mean \textit{MER} of 0.157, significantly higher than the present study, but this is likely due to differences in the data sets that were used. Graham and Roll used read speech recordings from the Speech Accent Archive (SAA) \citep{weinberger_speech_2015} and likely calculated \textit{MER} by comparing Whisper transcriptions to the original (phonologically difficult) paragraph that the speakers were asked to read. Many recordings in the SAA contain mistakes and disfluencies that are not present in the original paragraph. Another study by \citet{ballier_whisper_2024} used Whisper-large-v2 (an earlier model) to transcribe read speech from French learners and found a mean WER (not \textit{MER}) of 0.132, or 86.8\% accuracy. This study also calculated the error rate using the original prompt text rather than transcriptions of the recordings, so this score may also have been affected by speaker errors. The results of both of these studies highlight how critical accurate transcription data is for reliable ASR testing. Even though the L2-ARCTIC corpus used for the present study included human transcriptions for each recording, several mistakes and inconsistencies were identified which needed to be corrected. One other study \citep{wills_automatic_2023} had Whisper transcribe Dutch speech from non-native children and found a WER of 0.248, or 75.2\% accuracy, for read speech and 0.338, or 66.2\% accuracy, for dialogue speech.

A number of recent studies have looked at other commercial ASR systems and found lower levels of accuracy. \citet{mccrocklin_asr_2019} compared Windows Speech Recognition and Google Voice Typing on both read and spontaneous non-native speech. For read speech and spontaneous speech, accuracy rates were 74.44\% and 53.5\% for Windows Speech Recognition, and 88.61\% and 93.47\% for Google Voice Typing, respectively. \citet{hirai_speech--text_2024} looked at Google Voice Typing, Apple Dictation, Windows 10 Dictation, a web service called Dictation.io, and the iOS app Transcribe. English read speech recordings from 30 university students from seven different L1 backgrounds were transcribed through each of the five systems and found relatively poor accuracy: 64.37\% for Google, 48.91\% for Apple, 65.36\% for Windows 10, 57.91\% for Dictation.io, and 67.04\% for the Transcribe app. 

The present study sought to assess the accuracy of five of the newest, cutting-edge ASR systems on non-native accented learner English. As the results have clearly shown, different ASR systems excel at different tasks. For read speech, Whisper (mean \textit{MER} = 0.054) and AssemblyAI (mean \textit{MER} = 0.056) performed exceptionally well, reaching the human level accuracy level of reported by Amodei et al. (2015). However, when it came to spontaneous speech, they did not retain disfluencies as successfully as the other systems and did not perform as well. Nonetheless, these systems should be considered for read speech applications. RevAI was the slowest and performed the worst on read speech (mean \textit{MER} = 0.086), but achieved the best performance on the longer spontaneous speech recordings (mean \textit{MER} = 0.063), including transcription of disfluencies. This system may be ideal for transcribing longer-form recordings when processing time is not an issue. Deepgram was by far the fastest on all tasks, and while its accuracy received intermediate scores, it may be ideal when processing time is crucial, such as for interactive applications. Speechmatics, while still performing well, was the least notable system, with intermediate measures on all metrics. 

Instructors and researchers should take these differences into account when considering their specific use case. Language learning applications that handle both read speech (such as pronunciation practice) and spontaneous speech (such as open-ended questions) may want to consider different ASR systems for each task. However, when considering all the results from the present study, AssemblyAI performed consistently well across tasks and L1 backgrounds, making it perhaps the best overall single choice for different language learning contexts.

When looking at L1 background effects, US English speakers had dramatically lower error rates ($M = 0.007$) than all non-native groups, as expected. Among non-native groups, Hindi speakers showed lower error rates while Chinese and Vietnamese speakers showed higher error rates across both read and spontaneous speech. L1 background had a substantial impact ($F(6, 14) = 8.71$, $p < .001$, $\mathit{ges} = 0.778$) for read speech, which suggests that it might be necessary to adjust error thresholds based on typical ASR performance for specific L1 backgrounds. On the other hand, errors in ASR transcriptions of read speech can potentially indicate where additional pronunciation instruction and practice could be beneficial.

As for gender effects, female speakers showed lower error rates across both speech types, though not statistically significant. Gender differences varied considerably by L1 group, with the largest differences in Vietnamese (male $M = 0.181$; female $M = 0.105$) and Spanish speakers (male $M = 0.084$; female $M = 0.041$). This pattern of lower error rates for female speakers in English speech aligns with findings from several previous studies \citep{adda-decker_speech_2005,goldwater_which_2010,graham_evaluating_2024,koenecke_racial_2020}.

Longer processing time was likely due to API servers and did not correlate with accuracy (Spearman's $\rho = -0.002$ for read speech and $\rho = -0.087$ for spontaneous speech). Deepgram was consistently the most efficient (3.5x more efficient than RevAI for spontaneous speech) due to its speed.

\subsection{Whisper Initial Prompts and Hallucinations}
Even though Whisper did not have the lowest \textit{MER} on spontaneous speech, it saw the greatest improvement of any of the systems when retaining disfluencies (from 0.142 when omitted to 0.090 when retained). As mentioned earlier, Whisper does not have a simple option for retaining disfluencies, but instead an initial prompt can be used. While processing the data for this study, several different initial prompts were tested. What quickly became apparent was that using an initial prompt causes hallucinations (the inclusion words or expressions that are not in the recording) of varying severity. The subsequent process of trial and error may be of interest to other Whisper users.

The first initial prompt tested was simply ``um, uh'' in order to provide the expected filler words. This resulted in a large hallucination in speaker NCC's transcription, where the expression ``and they are not going to pick the right suitcase'' was inserted (it did not appear in the ground truth or the Whisper transcription without the initial prompt). The transcription of speaker HJK got stuck in a loop: ``because there was a woman's in a rush and he was like i'm going to go to the airport and she was like i'm going to go to the airport and he was like i'm going to go to the airport''. To experiment with this initial prompt, variations in punctuation were tried such as ``um uh'' (no comma in between), ``um. uh.'' (with periods), but these resulted in similar hallucinations, such as the speaker YKWK's repeating hallucination ``the man found a briefcase in his suitcase the man found a briefcase in his suitcase''.

Next, an initial prompt giving the specific instructions ``include filler words um uh'' was tested. This resulted in the most bizarre hallucinations: NCC now had the much longer ``and they are not sure if they are right or wrong and then they are forced to pick the wrong suitcase'' hallucination, and speaker ERMS's transcription ended with a completely fabricated YouTube-style outro ``and i think that's it for this video i hope you enjoyed it and i'll see you in the next one''.

Finally, the initial prompt that resulted in the best performance was ``man woman suitcase city um uh''. This was attempted after seeing a suggestion online to include several words that appear in the ground truth before adding filler words. While some small hallucinations remained, this initial prompt resulted in the lowest mean \textit{MER} across all recordings and was used for Whisper's disfluencies retained condition in the main analysis. It would technically be possible to make many attempts with different initial prompts and then select the best transcriptions from each, but that would require much more work than using any of the other ASR systems. 

What is particularly interesting is that even though using this initial prompt greatly improved the transcription \textit{MER}, it did not always improve the filler word retention, which can be seen in the example in~\ref{app:C}. Overall, Whisper had the lowest filler detection rate of 0.408 when using the initial prompt---up from only 0.025 without it. Ultimately, using the initial prompt improved Whisper's accuracy, though not necessarily in the way that was expected, and only through extensive trial and error. If verbatim transcription of filler words is important for your use case, then Whisper's performance might be frustrating. However, if you only want a more readable and fluent transcription of the primary message without disfluencies, then Whisper performs superbly.

To illustrate Whisper's performance with and without the initial prompt, we can look at the transcriptions of speaker HKK, which saw the greatest improvement in \textit{MER} between conditions: \textit{MER} = 0.238 without the initial prompt and \textit{MER} = 0.033 with the initial prompt:

\noindent\textit{Ground Truth:}
\begin{quote}
\small\ttfamily
um okay in the city i think um there are many skyscrapers and um the many cars uh not not so many cars on the road and in the corner of the building the two people are walking into the same directions and they're carrying the same carrier i think i mean the suitcase i think um because they did not see each other they just bumped into each other and um and especially for the man he dropped his glasses um after bumping into each other um they seem to have a headache because there is the uh stars going around their over their heads and after that they said to just the i'm sorry to each other and um pick up their own suitcases but and they just did uh went to their own direction i mean their destinations but after that when just they opened their um suitcases they found out it it's just they pick up the wrong suitcase because for the man in his suitcase they i mean he just he found out that the the red underwear was in the suitcase whereas the the woman found out that in her suitcase there was the um the yellow striped uh necktie in the suitcase so 
\end{quote}

\noindent\textit{Whisper Transcription, No Initial Prompt (MER = 0.238), differences underlined:}
\begin{quote}
\small\ttfamily
in the city i think there are many skyscrapers and not so many cars on the road and in the corner of the building the two people are walking into the same directions and they're carrying the same suitcase i think because they did not see each other they just bumped into each other and especially for the man he dropped his glasses after bumping into each other they seem to have a headache because there \uline{are} stars going over their heads and after that they said i'm sorry to each other and pick up their own suitcases and they just went to their own destinations but after that when \uline{they} opened their suitcases they found out \uline{that} they \uline{picked} up the wrong suitcase because for the man in his suitcase i mean he just found out that the red underwear was in the suitcase whereas the \uline{woman} found out that in her suitcase there was the yellow striped necktie in the suitcase
\end{quote}

\noindent\textit{Whisper Transcription, With Initial Prompt ``man woman suitcase city um uh'' (\textit{MER} = 0.033), differences underlined:}
\begin{quote}
\small\ttfamily
\uline{um okay} in the city i think \uline{um} there are many skyscrapers and \uline{um the many cars} not so many cars on the road and in the corner of the building the two people are walking into the same directions and they're carrying the same \uline{carrier i think i mean the} suitcase i think \uline{um} because they did not see each other they just bumped into each other \uline{and um} and especially for the man he dropped his glasses \uline{um} after bumping into each other \uline{um} they seem to have a headache because there \uline{is the uh} stars going \uline{around their} over their heads and after that they said \uline{to just the} i'm sorry to each other and \uline{um} pick up their own suitcases \uline{but} and they just \uline{uh} went to their own \uline{direction i mean the} destinations but after that when \uline{just he} opened their \uline{um} suitcases they found out \uline{it just} they \uline{pick} up the wrong suitcase because for the man in his suitcase \uline{they} i mean he just \uline{he} found out that the \uline{the} red underwear was in the suitcase whereas the the women found out that in her suitcase there was the um the yellow striped uh necktie in the suitcase so
\end{quote}

As can be seen, the default Whisper transcription with no initial prompt is much more fluent and readable, removing most disfluencies and even making grammatical adjustments such as changing ``there is the uh stars'' to ``there are stars''.

\subsection{Limitations}
This study investigated the performance of five cutting-edge ASR systems on non-native English using an available data set containing the speech of speakers from six different L1 backgrounds. While this can help paint a general picture of each system's performance, accuracy will vary by L1, as seen in this study's results. How each system will handle L1s other than those included in this study cannot be predicted with confidence, so they should be tested individually before choosing the best ASR system. Further studies might test other L1 backgrounds or look at performance on specific L1s or accents. 

As stated in the methods section, each ASR system was used in its default state, with the exception of adjusting disfluency retention settings which were tested and reported on. It is possible to fine tune ASR systems on your own training data, which will most likely lead to even better accuracy for your specific use case.

\subsection{Future directions}
As ASR systems have reached human-level accuracy on native speech, it is hoped that more and more researchers and developers turn their attention to the automatic transcription of non-native learner speech. It is quite evident that the most critical element for improving ASR transcription of non-native speech is training data. Creating high quality training data is expensive, which explains why many of the best systems are commercial. However, creating and openly sharing non-native speech data with high quality transcriptions can greatly benefit both ASR training and testing, as can fine-tuning ASR models and making them publicly available. There is a concerted push for releasing open source models---as seen with Whisper---which allows developers and researchers to freely modify and redistribute them. We have already seen a wide variety of Whisper-based variants emerge; the cloud-based machine learning platform Hugging Face (\url{https://huggingface.co}) lists over 10,000 Whisper-based models, though most are for improved speed and reduced computational requirements. Perhaps we will begin to see more Whisper models trained on non-native speech. If there is enough demand for ASR for non-native speech, commercial models may also begin to appear. There is still room for improvement in ASR, and I anticipate more advances in the coming years.

ASR is a rapidly developing field. This study used the newest models available at the time of writing, but new models are released frequently. Nevertheless, it is hoped that this study can provide some guidance to language educators and researchers who are interested in applying ASR to their work.

\section*{AI/LLM Disclosure Statement}
In preparation of this manuscript, the large language model \texttt{Claude-3-5-sonnet-20241022} (Anthropic) was used to assist in developing the Python and R scripts used for data analysis. The model was used solely for coding assistance, debugging, and standardizing statistical reporting formats, and did not contribute to the generation of any original ideas. All written content, including the broader manuscript text, analysis, interpretation, and conclusions, is original and was composed by the author without the use of AI assistance. The author alone is responsible for any inaccuracies present in the manuscript.

\printbibliography

@article{browne_new_2014,
	title = {The {New} {General} {Service} {List} {Version} 1.01: {Getting} {Better} {All} the {Time}},
	volume = {11},
	number = {1},
	journal = {Korea TESOL Journal},
	author = {Browne, Charles},
	year = {2014},
	pages = {35--50},
}

@inproceedings{morris_wer_2004,
	title = {From {WER} and {RIL} to {MER} and {WIL}: improved evaluation measures for connected speech recognition},
	shorttitle = {From {WER} and {RIL} to {MER} and {WIL}},
	url = {https://www.isca-speech.org/archive/interspeech_2004/morris04_interspeech.html},
	doi = {10.21437/Interspeech.2004-668},
	language = {en},
	urldate = {2022-12-29},
	booktitle = {Interspeech 2004},
	publisher = {ISCA},
	author = {Morris, Andrew Cameron and Maier, Viktoria and Green, Phil},
	month = oct,
	year = {2004},
	pages = {2765--2768},
}

@incollection{derwing_fluency_2022,
	address = {New York},
	edition = {1},
	title = {Fluency},
	isbn = {978-1-00-302249-7},
	url = {https://www.taylorfrancis.com/books/9781003022497/chapters/10.4324/9781003022497-17},
	language = {en},
	urldate = {2023-04-17},
	booktitle = {The {Routledge} {Handbook} of {Second} {Language} {Acquisition} and {Speaking}},
	publisher = {Routledge},
	author = {Kahng, Jimin},
	collaborator = {Derwing, Tracey M. and Munro, Murray J. and Thomson, Ron I.},
	month = jan,
	year = {2022},
	doi = {10.4324/9781003022497-17},
	pages = {188--200},
}

@incollection{radford_robust_2022,
	title = {Robust speech recognition via large-scale weak supervision},
	language = {en},
	booktitle = {Proceedings of {Machine} {Learning} {Research}},
	publisher = {PMLR},
	author = {Radford, Alec and Kim, Jong Wook and Xu, Tao and Brockman, Greg and McLeavey, Christine and Sutskever, Ilya},
	year = {2022},
	pages = {28492--28518},
}

@misc{vaessen_jiwer_2018,
	title = {jiwer},
	url = {https://github.com/jitsi/jiwer},
	author = {Vaessen, Nik},
	year = {2018},
}

@article{graham_evaluating_2024,
	title = {Evaluating {OpenAI}'s {Whisper} {ASR}: {Performance} analysis across diverse accents and speaker traits},
	volume = {4},
	issn = {2691-1191},
	shorttitle = {Evaluating {OpenAI}'s {Whisper} {ASR}},
	url = {https://pubs.aip.org/jel/article/4/2/025206/3267247/Evaluating-OpenAI-s-Whisper-ASR-Performance},
	doi = {10.1121/10.0024876},
	language = {en},
	number = {2},
	urldate = {2024-10-13},
	journal = {JASA Express Letters},
	author = {Graham, Calbert and Roll, Nathan},
	month = feb,
	year = {2024},
	pages = {025206},
}

@misc{weinberger_speech_2015,
	address = {George Mason University},
	title = {Speech accent archive},
	url = {http://accent.gmu.edu},
	author = {Weinberger, Steven},
	year = {2015},
}

@inproceedings{zhao_l2-arctic_2018,
	title = {L2-{ARCTIC}: {A} non-native {English} speech corpus},
	shorttitle = {L2-{ARCTIC}},
	url = {https://www.isca-archive.org/interspeech_2018/zhao18b_interspeech.html},
	doi = {10.21437/Interspeech.2018-1110},
	language = {en},
	urldate = {2024-12-19},
	booktitle = {Interspeech 2018},
	publisher = {ISCA},
	author = {Zhao, Guanlong and Sonsaat, Sinem and Silpachai, Alif and Lucic, Ivana and Chukharev-Hudilainen, Evgeny and Levis, John and Gutierrez-Osuna, Ricardo},
	month = sep,
	year = {2018},
	pages = {2783--2787},
}

@article{hirai_speech--text_2024,
	title = {Speech-to-text applications’ accuracy in {English} language learners’ speech transcription},
	volume = {28},
	url = {https://hdl.handle.net/10125/73555},
	language = {en},
	number = {1},
	journal = {Language Learning \& Technology},
	author = {Hirai, Akiyo and Kovalyova, Angelina},
	year = {2024},
	pages = {1--22},
}

@inproceedings{kominek_cmu_2004,
	address = {Pittsburgh, PA},
	title = {The {CMU} {ARCTIC} speech databases},
	url = {https://www.isca-archive.org/ssw_2004/kominek04b_ssw.pdf},
	language = {en},
	publisher = {International Speech Communication Association},
	author = {Kominek, John and Black, Alan W},
	year = {2004},
	pages = {223--224},
}

@article{thi-nhu_ngo_effectiveness_2024,
	title = {The effectiveness of automatic speech recognition in {ESL}/{EFL} pronunciation: {A} meta-analysis},
	volume = {36},
	issn = {0958-3440, 1474-0109},
	shorttitle = {The effectiveness of automatic speech recognition in {ESL}/{EFL} pronunciation},
	url = {https://www.cambridge.org/core/product/identifier/S0958344023000113/type/journal_article},
	doi = {10.1017/S0958344023000113},
	language = {en},
	number = {1},
	urldate = {2025-02-03},
	journal = {ReCALL},
	author = {Thi-Nhu Ngo, Thuy and Hao-Jan Chen, Howard and Kuo-Wei Lai, Kyle},
	month = jan,
	year = {2024},
	pages = {4--21},
}

@article{saito_effects_2012,
	title = {Effects of form‐focused instruction and corrective feedback on {L2} pronunciation development of /ɹ/ by {Japanese} learners of {English}},
	volume = {62},
	issn = {0023-8333, 1467-9922},
	url = {https://onlinelibrary.wiley.com/doi/10.1111/j.1467-9922.2011.00639.x},
	doi = {10.1111/j.1467-9922.2011.00639.x},
	language = {en},
	number = {2},
	urldate = {2025-02-03},
	journal = {Language Learning},
	author = {Saito, Kazuya},
	month = jun,
	year = {2012},
	pages = {595--633},
}

@misc{graves_sequence_2012,
	title = {Sequence transduction with recurrent neural networks},
	url = {http://arxiv.org/abs/1211.3711},
	doi = {10.48550/arXiv.1211.3711},
	language = {en},
	urldate = {2025-02-04},
	publisher = {arXiv},
	author = {Graves, Alex},
	month = nov,
	year = {2012},
	note = {arXiv:1211.3711 [cs]},
}

@misc{gulati_conformer_2020,
	title = {Conformer: {Convolution}-augmented transformer for speech recognition},
	shorttitle = {Conformer},
	url = {http://arxiv.org/abs/2005.08100},
	doi = {10.48550/arXiv.2005.08100},
	language = {en},
	urldate = {2025-02-04},
	publisher = {arXiv},
	author = {Gulati, Anmol and Qin, James and Chiu, Chung-Cheng and Parmar, Niki and Zhang, Yu and Yu, Jiahui and Han, Wei and Wang, Shibo and Zhang, Zhengdong and Wu, Yonghui and Pang, Ruoming},
	month = may,
	year = {2020},
	note = {arXiv:2005.08100 [eess]},
}

@misc{fox_introducing_2023,
	title = {Introducing {Nova}-2: {The} fastest, most accurate speech-to-text {API}},
	shorttitle = {Introducing nova-2},
	url = {https://deepgram.com/learn/nova-2-speech-to-text-api},
	language = {en},
	urldate = {2025-02-04},
	journal = {Deepgram},
	author = {Fox, Josh},
	month = sep,
	year = {2023},
}

@misc{jette_what_2024,
	title = {What makes {Rev}’s {V2} best in class?},
	shorttitle = {What makes rev’s v2 best in class?},
	url = {https://www.rev.com/blog/what-makes-revs-v2-best-in-class},
	language = {en},
	urldate = {2025-02-04},
	author = {Jette, Miguel},
	month = jan,
	year = {2024},
}

@misc{universal2_2024,
    author = {{"Universal-2: Comprehensive speech-to-text for solving last-mile challenges"}},
    title = {Universal-2: Comprehensive speech-to-text for solving last-mile challenges},
    organization = {AssemblyAI},
    url = {https://www.assemblyai.com/research/universal-2},
    year = {2024},
    month = oct,
    urldate = {2025-02-04}
}

@misc{steadman_ursa_2024,
	title = {Ursa 2: {Elevating} speech recognition across 50+ languages},
	shorttitle = {Ursa 2},
	url = {https://www.speechmatics.com/company/articles-and-news/ursa-2-elevating-speech-recognition-across-52-languages},
	language = {en},
	urldate = {2025-02-04},
	journal = {Speechmatics},
	author = {Steadman, Liam and Williams, Will},
	month = oct,
	year = {2024},
}

@incollection{lee_use_1996,
	address = {Boston, MA},
	title = {The use of recurrent neural networks in continuous speech recognition},
	volume = {355},
	isbn = {978-1-4612-8590-8},
	url = {http://link.springer.com/10.1007/978-1-4613-1367-0_10},
	language = {en},
	urldate = {2025-02-06},
	booktitle = {Automatic {Speech} and {Speaker} {Recognition}},
	publisher = {Springer US},
	author = {Robinson, Tony and Hochberg, Mike and Renals, Steve},
	editor = {Lee, Chin-Hui and Soong, Frank K. and Paliwal, Kuldip K.},
	year = {1996},
	doi = {10.1007/978-1-4613-1367-0_10},
	note = {Series Title: The Kluwer International Series in Engineering and Computer Science},
	pages = {233--258},
}

@article{robinson_recurrent_1991,
	title = {A recurrent error propagating network speech recognition system},
	volume = {1991},
	language = {en},
	number = {5},
	journal = {Computer Speech and Language},
	author = {Robinson, Tony and Fallside, Frank},
	year = {1991},
	pages = {259--274},
}

@inproceedings{saeki_intella_2024,
	address = {Kyoto, Japan},
	title = {{InteLLA}: {Intelligent} language learning assistant for assessing language proficiency through interviews and roleplays},
	shorttitle = {{InteLLA}},
	url = {https://aclanthology.org/2024.sigdial-1.34/},
	doi = {10.18653/v1/2024.sigdial-1.34},
	urldate = {2025-02-06},
	booktitle = {Proceedings of the 25th {Annual} {Meeting} of the {Special} {Interest} {Group} on {Discourse} and {Dialogue}},
	publisher = {Association for Computational Linguistics},
	author = {Saeki, Mao and Takatsu, Hiroaki and Kurata, Fuma and Suzuki, Shungo and Eguchi, Masaki and Matsuura, Ryuki and Takizawa, Kotaro and Yoshikawa, Sadahiro and Matsuyama, Yoichi},
	editor = {Kawahara, Tatsuya and Demberg, Vera and Ultes, Stefan and Inoue, Koji and Mehri, Shikib and Howcroft, David and Komatani, Kazunori},
	month = sep,
	year = {2024},
	pages = {385--399},
}

@misc{amann_augmenting_2024,
	title = {Augmenting automatic speech recognition models with disfluency detection},
	url = {http://arxiv.org/abs/2409.10177},
	doi = {10.48550/arXiv.2409.10177},
	language = {en},
	urldate = {2025-02-08},
	publisher = {arXiv},
	author = {Amann, Robin and Li, Zhaolin and Bruno, Barbara and Niehues, Jan},
	month = sep,
	year = {2024},
	note = {arXiv:2409.10177 [cs]},
}

@article{davis_automatic_1952,
	title = {Automatic recognition of spoken digits},
	volume = {24},
	issn = {0001-4966},
	url = {https://doi.org/10.1121/1.1906946},
	doi = {10.1121/1.1906946},
	number = {6},
	urldate = {2025-02-08},
	journal = {The Journal of the Acoustical Society of America},
	author = {Davis, K. H. and Biddulph, R. and Balashek, S.},
	month = nov,
	year = {1952},
	pages = {637--642},
}

@article{sakai_phonetic_1961,
	title = {Phonetic typewriter},
	volume = {33},
	issn = {0001-4966},
	url = {https://doi.org/10.1121/1.1936652},
	doi = {10.1121/1.1936652},
	number = {11\_Supplement},
	urldate = {2025-02-08},
	journal = {The Journal of the Acoustical Society of America},
	author = {Sakai, T. and Doshita, S.},
	month = nov,
	year = {1961},
	pages = {1664},
}

@article{lee_large-vocabulary_1988,
	series = {Word {Recognition} in {Large} {Vocabularies}},
	title = {On large-vocabulary speaker-independent continuous speech recognition},
	volume = {7},
	issn = {0167-6393},
	url = {https://www.sciencedirect.com/science/article/pii/0167639388900532},
	doi = {10.1016/0167-6393(88)90053-2},
	number = {4},
	urldate = {2025-02-08},
	journal = {Speech Communication},
	author = {Lee, Kai-Fu},
	month = dec,
	year = {1988},
	pages = {375--379},
}

@misc{cmusphinx_2025,
    author = {{CMUSphinx open source speech recognition}},
    title = {{CMUSphinx} open source speech recognition},
    organization = {CMUSphinx Open Source Speech Recognition},
    note = {\url{http://cmusphinx.github.io/}},
    year = {2025}
}

@article{rabiner_tutorial_1989,
	title = {A tutorial on hidden markov models and selected applications in speech recognition},
	volume = {77},
	doi = {10.1109/5.18626},
	language = {en},
	number = {2},
	journal = {Proceedings of the IEEE},
	author = {Rabiner, Lawrence R},
	year = {1989},
	pages = {257--286},
}

@article{dahl_context-dependent_2012,
	title = {Context-dependent pre-trained deep neural networks for large-vocabulary speech recognition},
	volume = {20},
	copyright = {https://ieeexplore.ieee.org/Xplorehelp/downloads/license-information/IEEE.html},
	issn = {1558-7916, 1558-7924},
	url = {http://ieeexplore.ieee.org/document/5740583/},
	doi = {10.1109/TASL.2011.2134090},
	language = {en},
	number = {1},
	urldate = {2025-02-08},
	journal = {IEEE Transactions on Audio, Speech, and Language Processing},
	author = {Dahl, G. E. and {Dong Yu} and {Li Deng} and Acero, A.},
	month = jan,
	year = {2012},
	pages = {30--42},
}

@article{hinton_deep_2012,
	title = {Deep neural networks for acoustic modeling in speech recognition: {The} shared views of four research groups},
	volume = {29},
	copyright = {https://ieeexplore.ieee.org/Xplorehelp/downloads/license-information/IEEE.html},
	issn = {1053-5888},
	shorttitle = {Deep neural networks for acoustic modeling in speech recognition},
	url = {http://ieeexplore.ieee.org/document/6296526/},
	doi = {10.1109/MSP.2012.2205597},
	language = {en},
	number = {6},
	urldate = {2025-02-08},
	journal = {IEEE Signal Processing Magazine},
	author = {Hinton, Geoffrey and Deng, Li and Yu, Dong and Dahl, George and Mohamed, Abdel-rahman and Jaitly, Navdeep and Senior, Andrew and Vanhoucke, Vincent and Nguyen, Patrick and Sainath, Tara and Kingsbury, Brian},
	month = nov,
	year = {2012},
	pages = {82--97},
}

@inproceedings{vaswani_attention_2017,
	address = {Long Beach, CA, USA},
	title = {Attention is all you need},
	url = {http://arxiv.org/abs/1706.03762},
	doi = {10.48550/arXiv.1706.03762},
	urldate = {2025-02-08},
	booktitle = {31st {Conference} on {Neural} {Information} {Processing} {Systems}},
	publisher = {arXiv},
	author = {Vaswani, Ashish and Shazeer, Noam and Parmar, Niki and Uszkoreit, Jakob and Jones, Llion and Gomez, Aidan N. and Kaiser, Lukasz and Polosukhin, Illia},
	year = {2017},
	note = {arXiv:1706.03762 [cs]},
	pages = {261--272},
}

@misc{chorowski_end--end_2014,
	title = {End-to-end continuous speech recognition using attention-based recurrent {NN}: {First} results},
	shorttitle = {End-to-end continuous speech recognition using attention-based recurrent {NN}},
	url = {http://arxiv.org/abs/1412.1602},
	doi = {10.48550/arXiv.1412.1602},
	urldate = {2025-02-08},
	publisher = {arXiv},
	author = {Chorowski, Jan and Bahdanau, Dzmitry and Cho, Kyunghyun and Bengio, Yoshua},
	month = dec,
	year = {2014},
	note = {arXiv:1412.1602 [cs]},
}

@inproceedings{chan_listen_2016,
	title = {Listen, attend and spell: {A} neural network for large vocabulary conversational speech recognition},
	shorttitle = {Listen, attend and spell},
	url = {https://ieeexplore.ieee.org/abstract/document/7472621},
	doi = {10.1109/ICASSP.2016.7472621},
	urldate = {2025-02-09},
	booktitle = {2016 {IEEE} {International} {Conference} on {Acoustics}, {Speech} and {Signal} {Processing} ({ICASSP})},
	author = {Chan, William and Jaitly, Navdeep and Le, Quoc and Vinyals, Oriol},
	month = mar,
	year = {2016},
	note = {ISSN: 2379-190X},
	pages = {4960--4964},
}

@misc{radford_improving_2018,
	title = {Improving language understanding by generative pre-training},
	url = {https://cdn.openai.com/research-covers/language-unsupervised/language_understanding_paper.pdf},
	language = {en},
	urldate = {2025-02-09},
	publisher = {OpenAI},
	author = {Radford, Alec and Narasimhan, Karthik and Salimans, Tim and Sutskever, Ilya},
	year = {2018},
}

@inproceedings{jaitly_application_2012,
	title = {Application of pretrained deep neural networks to large vocabulary speech recognition},
	url = {https://www.isca-archive.org/interspeech_2012/jaitly12_interspeech.html},
	doi = {10.21437/Interspeech.2012-10},
	language = {en},
	urldate = {2025-02-09},
	booktitle = {Interspeech 2012},
	publisher = {ISCA},
	author = {Jaitly, Navdeep and Nguyen, Patrick and Senior, Andrew and Vanhoucke, Vincent},
	month = sep,
	year = {2012},
	pages = {2578--2581},
}

@inproceedings{lin_study_2009,
	address = {Taipei, Taiwan},
	title = {A study on multilingual acoustic modeling for large vocabulary {ASR}},
	isbn = {978-1-4244-2353-8},
	url = {http://ieeexplore.ieee.org/document/4960588/},
	doi = {10.1109/ICASSP.2009.4960588},
	language = {en},
	urldate = {2025-02-09},
	booktitle = {2009 {IEEE} {International} {Conference} on {Acoustics}, {Speech} and {Signal} {Processing}},
	publisher = {IEEE},
	author = {Lin, Hui and Deng, Li and Yu, Dong and Gong, Yi-fan and Acero, Alex and Lee, Chin-Hui},
	month = apr,
	year = {2009},
	pages = {4333--4336},
}

@inproceedings{byrne_towards_2000,
	address = {Istanbul, Turkey},
	title = {Towards language independent acoustic modeling},
	volume = {2},
	isbn = {978-0-7803-6293-2},
	url = {http://ieeexplore.ieee.org/document/859138/},
	doi = {10.1109/ICASSP.2000.859138},
	language = {en},
	urldate = {2025-02-09},
	booktitle = {2000 {IEEE} {International} {Conference} on {Acoustics}, {Speech}, and {Signal} {Processing}. {Proceedings} ({Cat}. {No}.{00CH37100})},
	publisher = {IEEE},
	author = {Byrne, W. and Beyerlein, P. and Huerta, J.M. and Khudanpur, S. and Marthi, B. and Morgan, J. and Peterek, N. and Picone, J. and Vergyri, D. and Wang, T.},
	year = {2000},
	pages = {II1029--II1032},
}

@misc{bahdanau_neural_2014,
	title = {Neural machine translation by jointly learning to align and translate},
	url = {http://arxiv.org/abs/1409.0473},
	doi = {10.48550/arXiv.1409.0473},
	language = {en},
	urldate = {2025-02-09},
	publisher = {arXiv},
	author = {Bahdanau, Dzmitry and Cho, Kyunghyun and Bengio, Yoshua},
	year = {2014},
	note = {arXiv:1409.0473 [cs]},
}

@inproceedings{graves_connectionist_2006,
	address = {New York, NY, USA},
	series = {{ICML} '06},
	title = {Connectionist temporal classification: {Labelling} unsegmented sequence data with recurrent neural networks},
	isbn = {978-1-59593-383-6},
	shorttitle = {Connectionist temporal classification},
	url = {https://doi.org/10.1145/1143844.1143891},
	doi = {10.1145/1143844.1143891},
	urldate = {2025-02-09},
	booktitle = {Proceedings of the 23rd international conference on {Machine} learning},
	publisher = {Association for Computing Machinery},
	author = {Graves, Alex and Fernández, Santiago and Gomez, Faustino and Schmidhuber, Jürgen},
	month = jun,
	year = {2006},
	pages = {369--376},
}

@misc{r_core_team_r_2024,
	address = {Vienna, Austria},
	title = {R: {A} language and environment for statistical computing},
	url = {https://www.R-project.org/},
	publisher = {R Foundation for Statistical Computing},
	author = {R Core Team},
	year = {2024},
}

@misc{rstudio_team_rstudio_2024,
	address = {Boston, MA},
	title = {{RStudio}: integrated development environment for {R}},
	url = {https://posit.co/products/open-source/rstudio/},
	publisher = {Posit Software, PBC},
	author = {RStudio Team},
	year = {2024},
}

@misc{lawrence_ez_2016,
	title = {ez: {Easy} analysis and visualization of factorial experiments},
	url = {https://CRAN.R-project.org/package=ez},
	author = {Lawrence, Michael A.},
	month = nov,
	year = {2016},
}

@misc{amodei_deep_2015,
	title = {Deep {Speech} 2: {End}-to-end speech recognition in {English} and {Mandarin}},
	shorttitle = {Deep speech 2},
	url = {http://arxiv.org/abs/1512.02595},
	doi = {10.48550/arXiv.1512.02595},
	urldate = {2025-02-11},
	publisher = {arXiv},
	author = {Amodei, Dario and Anubhai, Rishita and Battenberg, Eric and Case, Carl and Casper, Jared and Catanzaro, Bryan and Chen, Jingdong and Chrzanowski, Mike and Coates, Adam and Diamos, Greg and Elsen, Erich and Engel, Jesse and Fan, Linxi and Fougner, Christopher and Han, Tony and Hannun, Awni and Jun, Billy and LeGresley, Patrick and Lin, Libby and Narang, Sharan and Ng, Andrew and Ozair, Sherjil and Prenger, Ryan and Raiman, Jonathan and Satheesh, Sanjeev and Seetapun, David and Sengupta, Shubho and Wang, Yi and Wang, Zhiqian and Wang, Chong and Xiao, Bo and Yogatama, Dani and Zhan, Jun and Zhu, Zhenyao},
	month = dec,
	year = {2015},
	note = {arXiv:1512.02595 [cs]},
}

@article{bell_adaptation_2021,
	title = {Adaptation algorithms for neural network-based speech recognition: an overview},
	volume = {2},
	issn = {2644-1322},
	shorttitle = {Adaptation algorithms for neural network-based speech recognition},
	url = {http://arxiv.org/abs/2008.06580},
	doi = {10.1109/OJSP.2020.3045349},
	urldate = {2025-02-12},
	journal = {IEEE Open Journal of Signal Processing},
	author = {Bell, Peter and Fainberg, Joachim and Klejch, Ondrej and Li, Jinyu and Renals, Steve and Swietojanski, Pawel},
	year = {2021},
	note = {arXiv:2008.06580 [eess]},
	pages = {33--66},
}

@article{koenecke_racial_2020,
	title = {Racial disparities in automated speech recognition},
	volume = {117},
	issn = {0027-8424, 1091-6490},
	url = {https://pnas.org/doi/full/10.1073/pnas.1915768117},
	doi = {10.1073/pnas.1915768117},
	language = {en},
	number = {14},
	urldate = {2025-02-12},
	journal = {Proceedings of the National Academy of Sciences},
	author = {Koenecke, Allison and Nam, Andrew and Lake, Emily and Nudell, Joe and Quartey, Minnie and Mengesha, Zion and Toups, Connor and Rickford, John R. and Jurafsky, Dan and Goel, Sharad},
	month = apr,
	year = {2020},
	pages = {7684--7689},
}

@article{coniam_voice_1999,
	title = {Voice recognition software accuracy with second language speakers of {English}},
	volume = {27},
	copyright = {https://www.elsevier.com/tdm/userlicense/1.0/},
	issn = {0346251X},
	url = {https://linkinghub.elsevier.com/retrieve/pii/S0346251X98000499},
	doi = {10.1016/S0346-251X(98)00049-9},
	language = {en},
	number = {1},
	urldate = {2025-02-12},
	journal = {System},
	author = {Coniam, D.},
	month = mar,
	year = {1999},
	pages = {49--64},
}

@article{derwing_does_2000,
	title = {Does popular speech recognition software work with {ESL} speech?},
	volume = {34},
	issn = {0039-8322},
	url = {https://www.jstor.org/stable/3587748},
	doi = {10.2307/3587748},
	number = {3},
	urldate = {2025-02-12},
	journal = {TESOL Quarterly},
	author = {Derwing, Tracey M. and Munro, Murray J. and Carbonaro, Michael},
	year = {2000},
	note = {Publisher: [Wiley, Teachers of English to Speakers of Other Languages, Inc. (TESOL)]},
	pages = {592--603},
}

@inproceedings{vergyri_automatic_2010,
	title = {Automatic speech recognition of multiple accented {English} data},
	url = {https://www.isca-archive.org/interspeech_2010/vergyri10_interspeech.html},
	doi = {10.21437/Interspeech.2010-477},
	urldate = {2025-02-12},
	author = {Vergyri, Dimitra and Lamel, Lori and Gauvain, Jean-Luc},
	year = {2010},
	pages = {1652--1655},
}

@article{ehsani_speech_1998,
	title = {Speech technology in computer-aided language},
	volume = {2},
	url = {http://llt.msu.edu/vol2num1/article3/},
	language = {en},
	number = {1},
	journal = {Language Learning \& Technology},
	author = {Ehsani, Farzad and Knodt, Eva},
	year = {1998},
	pages = {54--73},
}

@article{derwing_relationship_2009,
	title = {The relationship between {L1} fluency and {L2} fluency development},
	volume = {31},
	issn = {1470-1545},
	doi = {10.1017/S0272263109990015},
	number = {4},
	journal = {Studies in Second Language Acquisition},
	author = {Derwing, Tracey M. and Munro, Murray J. and Thomson, Ronald I. and Rossiter, Marian J.},
	year = {2009},
	note = {Place: United Kingdom
Publisher: Cambridge University Press},
	pages = {533--557},
}

@inproceedings{mccrocklin_asr_2019,
	address = {Ames, IA},
	title = {{ASR} dictation program accuracy: {Have} current programs improved?},
	language = {en},
	booktitle = {Proceedings of the 10th {Pronunciation} in {Second} {Language} {Learning} and {Teaching} {Conference}},
	author = {McCrocklin, Shannon and Humaidan, Abdulsamad and Edalatishams, Idée},
	editor = {Levis, J. and Nagle, C. and Todey, E.},
	year = {2019},
}

@article{wills_automatic_2023,
	title = {Automatic speech recognition of non-native child speech for language learning applications},
	volume = {113},
	issn = {2190-6807},
	url = {http://arxiv.org/abs/2306.16710},
	doi = {10.4230/OASIcs.SLATE.2023.7},
	urldate = {2025-02-18},
	journal = {OASIcs, Volume 113, SLATE 2023},
	author = {Wills, Simone and Bai, Yu and Tejedor-Garcia, Cristian and Cucchiarini, Catia and Strik, Helmer},
	year = {2023},
	note = {arXiv:2306.16710 [cs]},
	pages = {7:1--7:8},
}

@article{ballier_whisper_2024,
	title = {Whisper for {L2} speech scoring},
	volume = {27},
	issn = {1572-8110},
	url = {https://doi.org/10.1007/s10772-024-10141-5},
	doi = {10.1007/s10772-024-10141-5},
	language = {en},
	number = {4},
	urldate = {2025-02-18},
	journal = {International Journal of Speech Technology},
	author = {Ballier, Nicolas and Arnold, Taylor and Méli, Adrien and Thurston, Tori and Yunès, Jean-Baptiste},
	month = dec,
	year = {2024},
	pages = {923--934},
}

@inproceedings{stolcke_comparing_2017,
	title = {Comparing human and machine errors in conversational speech transcription},
	url = {http://arxiv.org/abs/1708.08615},
	doi = {10.21437/Interspeech.2017-1544},
	urldate = {2025-02-20},
	booktitle = {Interspeech 2017},
	author = {Stolcke, Andreas and Droppo, Jasha},
	month = aug,
	year = {2017},
	note = {arXiv:1708.08615 [cs]},
	pages = {137--141},
}

@inproceedings{panayotov_librispeech_2015,
	address = {South Brisbane, Queensland, Australia},
	title = {Librispeech: {An} {ASR} corpus based on public domain audio books},
	isbn = {978-1-4673-6997-8},
	shorttitle = {Librispeech},
	url = {http://ieeexplore.ieee.org/document/7178964/},
	doi = {10.1109/ICASSP.2015.7178964},
	language = {en},
	urldate = {2025-02-20},
	booktitle = {2015 {IEEE} {International} {Conference} on {Acoustics}, {Speech} and {Signal} {Processing} ({ICASSP})},
	publisher = {IEEE},
	author = {Panayotov, Vassil and Chen, Guoguo and Povey, Daniel and Khudanpur, Sanjeev},
	month = apr,
	year = {2015},
	pages = {5206--5210},
}

@misc{przybocki_2000_2001,
	title = {2000 {NIST} speaker recognition evaluation},
	url = {https://catalog.ldc.upenn.edu/LDC2001S97},
	doi = {10.35111/EX24-J205},
	urldate = {2025-02-20},
	publisher = {Linguistic Data Consortium},
	author = {{Przybocki} and {Martin}},
	year = {2001},
	note = {Artwork Size: 2735785 KB
Pages: 2735785 KB},
}

@inproceedings{adda-decker_speech_2005,
	title = {Do speech recognizers prefer female speakers?},
	url = {https://www.isca-archive.org/interspeech_2005/addadecker05_interspeech.html},
	doi = {10.21437/Interspeech.2005-699},
	language = {en},
	urldate = {2025-02-23},
	booktitle = {Interspeech 2005},
	publisher = {ISCA},
	author = {Adda-Decker, Martine and Lamel, Lori},
	month = sep,
	year = {2005},
	pages = {2205--2208},
}

@article{goldwater_which_2010,
	title = {Which words are hard to recognize? {Prosodic}, lexical, and disfluency factors that increase speech recognition error rates},
	volume = {52},
	issn = {0167-6393},
	shorttitle = {Which words are hard to recognize?},
	url = {https://www.sciencedirect.com/science/article/pii/S0167639309001599},
	doi = {10.1016/j.specom.2009.10.001},
	number = {3},
	urldate = {2025-02-23},
	journal = {Speech Communication},
	author = {Goldwater, Sharon and Jurafsky, Dan and Manning, Christopher D.},
	month = mar,
	year = {2010},
	pages = {181--200},
}

@article{mcguire_assessing_2025,
	title = {Assessing {Whisper} automatic speech recognition and {WER} scoring for elicited imitation: {Steps} toward automation},
	volume = {4},
	doi = {https://doi.org/10.1016/j.rmal.2025.100197},
	language = {en},
	number = {1},
	journal = {Research Methods in Applied Linguistics},
	author = {McGuire, Michael and Larson-Hall, Jenifer},
	year = {2025},
}

@article{mccrocklin_asr-based_2019,
	title = {{ASR}-based dictation practice for second language pronunciation improvement},
	volume = {5},
	issn = {2215-1931, 2215-194X},
	url = {http://www.jbe-platform.com/content/journals/10.1075/jslp.16034.mcc},
	doi = {10.1075/jslp.16034.mcc},
	language = {en},
	number = {1},
	urldate = {2025-03-28},
	journal = {Journal of Second Language Pronunciation},
	author = {McCrocklin, Shannon},
	month = mar,
	year = {2019},
	pages = {98--118},
}

@article{suzuki_recognition_1961,
	title = {Recognition of {Japanese} vowels},
	volume = {8},
	number = {37},
	journal = {Journal of the Radio Research Laboratory},
	author = {Suzuki, Joji},
	year = {1961},
	note = {Publisher: Tokyo: Radio Research Laboratory},
}

@article{shadiev_review_2023,
	title = {Review of research on applications of speech recognition technology to assist language learning},
	volume = {35},
	issn = {0958-3440, 1474-0109},
	url = {https://www.cambridge.org/core/product/identifier/S095834402200012X/type/journal_article},
	doi = {10.1017/S095834402200012X},
	language = {en},
	number = {1},
	urldate = {2024-06-10},
	journal = {ReCALL},
	author = {Shadiev, Rustam and Liu, Jiawen},
	month = jan,
	year = {2023},
	pages = {74--88},
}

@inproceedings{neri_automatic_2003,
	address = {Barcelona, Spain},
	title = {Automatic speech recognition for second language learning: {How} and why it actually works},
	url = {http://www.internationalphoneticassociation.org/icphs/icphs2003},
	language = {en},
	publisher = {International Phonetic Association International Phonetic Association},
	author = {Neri, Ambra and Cucchiarini, Catia and Strik, Wilhelmus},
	year = {2003},
	pages = {1157--1160},
}

@misc{hinsvark_accented_2021,
	title = {Accented speech recognition: {A} survey},
	shorttitle = {Accented speech recognition},
	url = {http://arxiv.org/abs/2104.10747},
	doi = {10.48550/arXiv.2104.10747},
	urldate = {2025-02-12},
	publisher = {arXiv},
	author = {Hinsvark, Arthur and Delworth, Natalie and Rio, Miguel Del and McNamara, Quinten and Dong, Joshua and Westerman, Ryan and Huang, Michelle and Palakapilly, Joseph and Drexler, Jennifer and Pirkin, Ilya and Bhandari, Nishchal and Jette, Miguel},
	month = jun,
	year = {2021},
	note = {arXiv:2104.10747 [cs]},
}

@misc{do_improving_2024,
	title = {Improving accented speech recognition using data augmentation based on unsupervised text-to-speech synthesis},
	url = {http://arxiv.org/abs/2407.04047},
	doi = {10.48550/arXiv.2407.04047},
	urldate = {2025-02-11},
	publisher = {arXiv},
	author = {Do, Cong-Thanh and Imai, Shuhei and Doddipatla, Rama and Hain, Thomas},
	month = jul,
	year = {2024},
	note = {arXiv:2407.04047 [cs]},
}

@misc{xiong_achieving_2017,
	title = {Achieving human parity in conversational speech recognition},
	url = {http://arxiv.org/abs/1610.05256},
	doi = {10.48550/arXiv.1610.05256},
	urldate = {2025-02-20},
	publisher = {arXiv},
	author = {Xiong, W. and Droppo, J. and Huang, X. and Seide, F. and Seltzer, M. and Stolcke, A. and Yu, D. and Zweig, G.},
	month = feb,
	year = {2017},
	note = {arXiv:1610.05256 [cs]},
}

@article{fiscus_2000_2000,
	title = {2000 {NIST} evaluation of conversational speech recognition over the telephone: {English} and {Mandarin} performance results},
	language = {en},
	author = {Fiscus, Jonathan G and Fisher, William M and Martin, Alvin F and Przybocki, Mark A and Pallett, David S},
	year = {2000},
}

@incollection{redford_fluency_2015,
	edition = {1},
	title = {Fluency and disfluency},
	isbn = {978-0-470-65993-9},
	url = {https://onlinelibrary.wiley.com/doi/10.1002/9781118584156.ch20},
	language = {en},
	urldate = {2025-02-17},
	booktitle = {The {Handbook} of {Speech} {Production}},
	publisher = {Wiley},
	author = {Lickley, Robin J.},
	editor = {Redford, Melissa A.},
	month = mar,
	year = {2015},
	doi = {10.1002/9781118584156.ch20},
	pages = {445--474},
}

@misc{rev_press_rev_2022,
	title = {Rev improves accuracy by 30\% with new {ASR} model},
	url = {https://www.rev.com/blog/rev-improves-accuracy-by-over-25-with-launch-of-new-v2-asr-model},
	language = {en},
	urldate = {2025-02-06},
	author = {{Rev Press}},
	month = apr,
	year = {2022},
}
\appendix
\renewcommand{\thesection}{Appendix \Alph{section}}

\begin{landscape}
\section{Mean MER and SD by L1, Gender, Speaker, and ASR System}
\label{app:A}

\begingroup
\footnotesize  
\setlength{\tabcolsep}{5pt}
\renewcommand{\arraystretch}{1.2}
\begin{center}
\begin{tabular}{llllccccccccccc}
\toprule
& & & & \multicolumn{2}{c}{AssemblyAI} & \multicolumn{2}{c}{Deepgram} & \multicolumn{2}{c}{RevAI} & \multicolumn{2}{c}{Speechmatics} & \multicolumn{2}{c}{Whisper} \\
\cmidrule(r){5-6} \cmidrule(r){7-8} \cmidrule(r){9-10} \cmidrule(r){11-12} \cmidrule(r){13-14}
L1 & Gender & Speaker & $N$ & $M$ & $SD$ & $M$ & $SD$ & $M$ & $SD$ & $M$ & $SD$ & $M$ & $SD$ \\
\midrule
Arabic & F & SKA & 100 & 0.079 & 0.134 & 0.135 & 0.169 & 0.136 & 0.166 & 0.113 & 0.154 & 0.082 & 0.167 \\
& F & ZHAA & 100 & 0.046 & 0.099 & 0.062 & 0.120 & 0.060 & 0.104 & 0.042 & 0.087 & 0.036 & 0.085 \\
& M & ABA & 100 & 0.043 & 0.082 & 0.057 & 0.106 & 0.082 & 0.122 & 0.058 & 0.099 & 0.039 & 0.092 \\
& M & YBAA & 100 & 0.034 & 0.087 & 0.050 & 0.099 & 0.049 & 0.098 & 0.043 & 0.087 & 0.031 & 0.081 \\
\midrule
Chinese & F & LXC & 100 & 0.062 & 0.103 & 0.096 & 0.134 & 0.088 & 0.121 & 0.077 & 0.133 & 0.060 & 0.109 \\
& F & NCC & 100 & 0.047 & 0.085 & 0.084 & 0.142 & 0.081 & 0.115 & 0.073 & 0.116 & 0.036 & 0.074 \\
& M & BWC & 100 & 0.070 & 0.108 & 0.086 & 0.130 & 0.096 & 0.130 & 0.081 & 0.117 & 0.061 & 0.103 \\
& M & TXHC & 100 & 0.039 & 0.084 & 0.078 & 0.123 & 0.068 & 0.111 & 0.082 & 0.135 & 0.056 & 0.108 \\
\midrule
Hindi & F & SVBI & 100 & 0.012 & 0.054 & 0.035 & 0.103 & 0.031 & 0.085 & 0.023 & 0.056 & 0.016 & 0.060 \\
& F & TNI & 100 & 0.031 & 0.082 & 0.034 & 0.081 & 0.045 & 0.086 & 0.034 & 0.082 & 0.026 & 0.068 \\
& M & ASI & 100 & 0.018 & 0.055 & 0.030 & 0.069 & 0.041 & 0.082 & 0.031 & 0.063 & 0.016 & 0.052 \\
& M & RRBI & 100 & 0.020 & 0.068 & 0.037 & 0.083 & 0.038 & 0.077 & 0.038 & 0.079 & 0.025 & 0.071 \\
\midrule
Korean & F & HJK & 100 & 0.037 & 0.087 & 0.056 & 0.106 & 0.044 & 0.090 & 0.037 & 0.080 & 0.035 & 0.082 \\
& F & YDCK & 100 & 0.036 & 0.089 & 0.047 & 0.109 & 0.053 & 0.109 & 0.047 & 0.104 & 0.032 & 0.079 \\
& M & HKK & 100 & 0.068 & 0.093 & 0.093 & 0.117 & 0.107 & 0.137 & 0.109 & 0.149 & 0.052 & 0.089 \\
& M & YKWK & 100 & 0.026 & 0.076 & 0.035 & 0.085 & 0.040 & 0.085 & 0.040 & 0.086 & 0.014 & 0.046 \\
\midrule
Spanish & F & MBMPS & 100 & 0.031 & 0.069 & 0.038 & 0.077 & 0.049 & 0.101 & 0.032 & 0.062 & 0.025 & 0.056 \\
& F & NJS & 100 & 0.034 & 0.090 & 0.050 & 0.115 & 0.060 & 0.111 & 0.057 & 0.121 & 0.032 & 0.083 \\
& M & EBVS & 100 & 0.083 & 0.125 & 0.126 & 0.159 & 0.146 & 0.156 & 0.133 & 0.168 & 0.084 & 0.137 \\
& M & ERMS & 100 & 0.037 & 0.075 & 0.051 & 0.095 & 0.086 & 0.133 & 0.051 & 0.085 & 0.042 & 0.088 \\
\midrule
US English & F & CLB & 100 & 0.002 & 0.013 & 0.008 & 0.045 & 0.007 & 0.035 & 0.006 & 0.031 & 0.006 & 0.030 \\
& F & SLT & 100 & 0.007 & 0.034 & 0.008 & 0.043 & 0.006 & 0.034 & 0.005 & 0.029 & 0.011 & 0.058 \\
& M & BDL & 100 & 0.010 & 0.046 & 0.011 & 0.042 & 0.009 & 0.038 & 0.007 & 0.034 & 0.007 & 0.033 \\
& M & RMS & 100 & 0.007 & 0.033 & 0.006 & 0.031 & 0.008 & 0.034 & 0.006 & 0.031 & 0.006 & 0.032 \\
\midrule
Vietnamese & F & PNV & 100 & 0.041 & 0.081 & 0.053 & 0.093 & 0.066 & 0.111 & 0.042 & 0.082 & 0.048 & 0.127 \\
& F & THV & 100 & 0.135 & 0.194 & 0.174 & 0.209 & 0.188 & 0.195 & 0.163 & 0.212 & 0.140 & 0.210 \\
& M & HQTV & 100 & 0.160 & 0.186 & 0.216 & 0.207 & 0.198 & 0.190 & 0.191 & 0.205 & 0.159 & 0.173 \\
& M & TLV & 100 & 0.148 & 0.174 & 0.203 & 0.200 & 0.209 & 0.201 & 0.178 & 0.199 & 0.149 & 0.193 \\
\bottomrule
\end{tabular}
\end{center}
\endgroup
\end{landscape}

\begin{landscape}
\section{MER for Spontaneous Speech by L1, Gender, ASR Model, and Disfluency Condition}
\label{app:B}

\begingroup
\footnotesize
\setlength{\tabcolsep}{5pt}
\renewcommand{\arraystretch}{1.2}
\begin{center}
\begin{tabular}{lllcccccccccc}
\toprule
& & & \multicolumn{2}{c}{AssemblyAI} & \multicolumn{2}{c}{Deepgram} & \multicolumn{2}{c}{RevAI} & \multicolumn{2}{c}{Speechmatics} & \multicolumn{2}{c}{Whisper} \\
\cmidrule(r){4-5} \cmidrule(r){6-7} \cmidrule(r){8-9} \cmidrule(r){10-11} \cmidrule(r){12-13}
& & & \multicolumn{10}{c}{Disfluency Condition} \\
\cmidrule(r){4-13}
L1 & Gender & Speaker & Omitted & Retained & Omitted & Retained & Omitted & Retained & Omitted & Retained & Omitted & Retained \\
\midrule
Arabic & F & ZHAA & 0.103 & 0.084 & 0.120 & 0.110 & 0.154 & 0.085 & 0.111 & 0.051 & 0.120 & 0.068 \\
Arabic & M & ABA & 0.135 & 0.115 & 0.103 & 0.064 & 0.064 & 0.026 & 0.115 & 0.082 & 0.045 & 0.045 \\
Arabic & M & YBAA & 0.078 & 0.078 & 0.035 & 0.035 & 0.043 & 0.035 & 0.034 & 0.034 & 0.195 & 0.043 \\
Chinese & F & LXC & 0.200 & 0.144 & 0.154 & 0.107 & 0.205 & 0.144 & 0.224 & 0.184 & 0.229 & 0.116 \\
Chinese & F & NCC & 0.283 & 0.211 & 0.240 & 0.116 & 0.220 & 0.102 & 0.228 & 0.110 & 0.276 & 0.270 \\
Chinese & M & BWC & 0.098 & 0.098 & 0.056 & 0.056 & 0.049 & 0.049 & 0.070 & 0.063 & 0.063 & 0.042 \\
Chinese & M & TXHC & 0.157 & 0.127 & 0.142 & 0.134 & 0.105 & 0.075 & 0.112 & 0.082 & 0.163 & 0.097 \\
Hindi & F & SVBI & 0.000 & 0.013 & 0.000 & 0.000 & 0.027 & 0.027 & 0.000 & 0.014 & 0.000 & 0.000 \\
Hindi & F & TNI & 0.125 & 0.058 & 0.125 & 0.078 & 0.132 & 0.033 & 0.125 & 0.026 & 0.132 & 0.179 \\
Hindi & M & RRBI & 0.070 & 0.070 & 0.078 & 0.061 & 0.079 & 0.061 & 0.095 & 0.095 & 0.070 & 0.070 \\
Korean & F & HJK & 0.075 & 0.043 & 0.062 & 0.031 & 0.056 & 0.025 & 0.062 & 0.025 & 0.106 & 0.037 \\
Korean & F & YDCK & 0.140 & 0.104 & 0.115 & 0.072 & 0.099 & 0.051 & 0.122 & 0.077 & 0.179 & 0.110 \\
Korean & M & HKK & 0.133 & 0.071 & 0.118 & 0.085 & 0.124 & 0.071 & 0.100 & 0.043 & 0.238 & 0.033 \\
Korean & M & YKWK & 0.181 & 0.124 & 0.135 & 0.090 & 0.117 & 0.067 & 0.119 & 0.051 & 0.215 & 0.099 \\
Spanish & F & MBMPS & 0.017 & 0.025 & 0.042 & 0.042 & 0.033 & 0.033 & 0.017 & 0.017 & 0.033 & 0.025 \\
Spanish & F & NJS & 0.055 & 0.054 & 0.047 & 0.023 & 0.047 & 0.031 & 0.039 & 0.023 & 0.077 & 0.062 \\
Spanish & M & EBVS & 0.168 & 0.178 & 0.146 & 0.132 & 0.085 & 0.070 & 0.146 & 0.139 & 0.189 & 0.105 \\
Spanish & M & ERMS & 0.062 & 0.076 & 0.085 & 0.100 & 0.070 & 0.062 & 0.055 & 0.047 & 0.093 & 0.092 \\
Vietnamese & F & PNV & 0.178 & 0.083 & 0.169 & 0.085 & 0.169 & 0.034 & 0.178 & 0.042 & 0.178 & 0.110 \\
Vietnamese & F & THV & 0.123 & 0.123 & 0.195 & 0.195 & 0.110 & 0.110 & 0.162 & 0.162 & 0.158 & 0.158 \\
Vietnamese & M & HQTV & 0.080 & 0.060 & 0.160 & 0.140 & 0.100 & 0.080 & 0.098 & 0.098 & 0.080 & 0.061 \\
Vietnamese & M & TLV & 0.228 & 0.171 & 0.171 & 0.122 & 0.179 & 0.122 & 0.252 & 0.195 & 0.280 & 0.161 \\
\midrule
Mean & & & 0.122 & 0.096 & 0.114 & 0.085 & 0.103 & 0.063 & 0.112 & 0.075 & 0.142 & 0.090 \\
\bottomrule
\end{tabular}
\end{center}
\endgroup
\end{landscape}
\begin{landscape}
\section{Transcriptions and Disfluencies for Speaker NCC -- L1 Chinese Female}
\label{app:C}
\begingroup
\scriptsize
\setlength{\tabcolsep}{3pt}
\renewcommand{\arraystretch}{1.0}
\begin{center}
\begin{tabular}{p{1.5cm}p{60em}ccp{10em}}
\toprule
Transcription & \multicolumn{1}{c}{Speaker NCC} & \textcolor{red}{Fillers} & \uline{Repetitions} & \multicolumn{1}{c}{\uuline{Revisions} (\textit{MER})} \\
\midrule
Ground truth & {\scriptsize\ttfamily \textcolor{red}{um} in a very beautiful \textcolor{red}{um} big city \textcolor{red}{uh} one woman and \textcolor{red}{uh} one man \textcolor{red}{uh} \uuline{which \textcolor{red}{uh} who} they both held the green suitcase met at \textcolor{red}{uh} one corner but certainly \uuline{they fell they crash} at each other and \textcolor{red}{uh} they blurred \uline{and \textcolor{red}{uh} and \textcolor{red}{uh}} although they \textcolor{red}{uh} \uline{stand stand} up immediately and say sorry to each other and \textcolor{red}{uh} they grab their suitcase and \textcolor{red}{uh} continue on their way however when they \textcolor{red}{uh} go back to their \uline{own own} house and open the suitcase \uuline{the man think that \textcolor{red}{uh} the man find that} they \textcolor{red}{uh} picked the wrong suitcase because in his case there is one red dress and that \textcolor{red}{uh} for the woman she \textcolor{red}{uh} her suitcase \textcolor{red}{uh} was \textcolor{red}{uh} yellow tie yeah} & \textcolor{red}{20} & 3 & which uh who; they fell they crash; the man think that uh the man find that \\
\midrule
AssemblyAI & {\scriptsize\ttfamily \textcolor{red}{um} in a \textcolor{red}{uh} very beautiful big city \textcolor{red}{uh} one woman and \textcolor{red}{uh} one man \textcolor{red}{uh} \uuline{who} they both held the green suitcase met had one corner but certainly \uuline{they felt they crashed} at each other and they burned and \textcolor{red}{um} \textcolor{red}{uh} although they \textcolor{red}{uh} stand up immediately and say sorry to each other and grab their suitcase and continue \textcolor{red}{uh} their way however when they \textcolor{red}{uh} go back to their own house and open the suitcase \uuline{the men think that \textcolor{red}{uh} the men find that} they \textcolor{red}{um} pick the suitcase because in his case there is one red dress and but \textcolor{red}{uh} for the woman her suitcase \textcolor{red}{uh} was a yellow tie yeah} & \textcolor{red}{14} & 0 & who (0.667); they felt they crashed (0.5); the men think that uh the men find that (0.222) \\
\midrule
Deepgram & {\scriptsize\ttfamily \textcolor{red}{um} in a very beautiful \textcolor{red}{um} big city \textcolor{red}{uh} 1 woman and \textcolor{red}{uh} 1 man \textcolor{red}{uh} \uuline{which \textcolor{red}{uh} who} they both held the green suitcase and met at one corner but certainly \uuline{they felt they crashed} at each other and \textcolor{red}{uh} they burned \uline{and \textcolor{red}{uh} and \textcolor{red}{uh}} although they \textcolor{red}{uh} \uline{stand stand} up \uline{immediately immediately} and say sorry to each other and \textcolor{red}{uh} they grab their suitcase and \textcolor{red}{uh} continue their way however when they \textcolor{red}{uh} go back to their \uline{own own} house and open the suitcase \uuline{the men think that \textcolor{red}{uh} the men find that} they \textcolor{red}{um} pick the wrong suitcase because in his case there is one red dress and that \textcolor{red}{uh} for the woman she her suitcase \textcolor{red}{uh} was a yellow tie yeah} & \textcolor{red}{17} & 4 & which uh who (0); they felt they crashed (0.5); the men think that uh the men find that (0.222) \\
\midrule
RevAI & {\scriptsize\ttfamily \textcolor{red}{um} in a very beautiful \textcolor{red}{um} big city \textcolor{red}{uh} one woman and \textcolor{red}{uh} one man \textcolor{red}{uh} \uuline{which \textcolor{red}{uh} who} they both held the green suitcase matt had \textcolor{red}{uh} one corner but certainly \uuline{they felt they crashed} at each other and they burned and \textcolor{red}{um} and \textcolor{red}{uh} although they \textcolor{red}{uh} \uline{stand stand} up immediately and say sorry to each other and they grab their suitcase and continue their way however when they \textcolor{red}{uh} go back to their \uline{own own} house and open the suitcase \uuline{the man think that \textcolor{red}{uh} the man find that} they \textcolor{red}{um} pick the wrong suitcase because in his case there is one red dress and but \textcolor{red}{uh} for the woman she \textcolor{red}{uh} her suitcase \textcolor{red}{uh} was \textcolor{red}{uh} yellow tie yeah} & \textcolor{red}{17} & 2 & which uh who (0); they felt they crashed (0.5); the man think that uh the man find that (0) \\
\midrule
Speechmatics & {\scriptsize\ttfamily \textcolor{red}{um} in a very beautiful \textcolor{red}{um} big city \textcolor{red}{uh} one woman and \textcolor{red}{uh} one man \textcolor{red}{uh} \uuline{which \textcolor{red}{uh} who} they both held the green suitcase and matt had one corner but certainly \uuline{they felt they crash} at each other and they blurred and \textcolor{red}{uh} and although they \textcolor{red}{uh} \uline{stand stand} up immediately and say sorry to each other and they grab their suitcase and \textcolor{red}{uh} continue their way however when they \textcolor{red}{uh} go back to their \uline{own own} house and open the suitcase \uuline{the men think that \textcolor{red}{uh} the men find that} they \textcolor{red}{uh} picked the wrong suitcase because in his case there is one red dress and that \textcolor{red}{uh} for the women she her suitcase \textcolor{red}{uh} was a yellow tie} & \textcolor{red}{14} & 2 & which uh who (0); they felt they crash (0.25); the men think that uh the men find that (0.222) \\
\midrule
Whisper & {\scriptsize\ttfamily \textcolor{red}{um} in a very beautiful \textcolor{red}{um} big city one woman and one man \uuline{which who} they both held the green suitcase met at one corner but certainly \uuline{they felt they crashed} at each other and they burned \uline{and and} although they \uline{stand stand} up immediately and say sorry to each other and they grab their suitcase and continue their way however when they go back to their \uline{own own} house and open the suitcase \uuline{the men think that the men find that} they pick the wrong suitcase and they are not going to pick the right suitcase because in his case there is one red dress and that for the women she her suitcase was a yellow tie} & \textcolor{red}{2} & 3 & which who (0.333); they felt they crashed (0.5); the men think that the men find that (0.333) \\
\bottomrule
\end{tabular}
\end{center}
\endgroup
\end{landscape}
    
    \end{document}